\documentclass{article}

\usepackage[eandd, final]{neurips_2026}

\usepackage[utf8]{inputenc} 
\usepackage[T1]{fontenc}    
\usepackage{hyperref}       
\usepackage{url}            
\usepackage{booktabs}       
\usepackage{amsfonts}       
\usepackage{nicefrac}       
\usepackage{microtype}      
\usepackage{xcolor}         
\usepackage{amsmath}
\usepackage{graphicx}
\usepackage{multirow}
\usepackage{enumitem}
\usepackage{adjustbox}
\usepackage{tcolorbox}
\usepackage{wrapfig}
\usepackage{cleveref}
\usepackage{caption}
\usepackage{array}
\newcolumntype{R}[1]{>{\raggedleft\arraybackslash}p{#1}}
\newcolumntype{L}[1]{>{\raggedright\arraybackslash}p{#1}}
\definecolor{mygrey}{HTML}{3a3a3a}

\title{Probing the Trajectories of Reasoning Traces \\ in Large Language Models}

\author{%
  Marthe Ballon\thanks{Corresponding author: marthe.ballon@vub.be} \\
  Data Analytics Laboratory\\
  Vrije Universiteit Brussel\\
  \texttt{Marthe.Ballon@vub.be} \\
\And
Brecht Verbeken \\
Data Analytics Laboratory\\
Vrije Universiteit Brussel\\
\texttt{Brecht.Verbeken@vub.be}\\
\AND
Vincent Ginis\\
Data Analyitcs Laboratory\\
Vrije Universiteit Brussel\\
SEAS, Harvard University\\
\texttt{Vincent.Ginis@vub.be}\\
\And Andres Algaba\\
Data Analytics Laboratory\\
Vrije Universiteit Brussel\\
\texttt{Andres.Algaba@vub.be}
}

\begin{document}

\maketitle

\begin{abstract}
Large language models (LLMs) increasingly solve difficult problems by producing reasoning traces before emitting a final response. However, it remains unclear how accuracy and decision commitment evolve along a reasoning trajectory, and whether intermediate trace segments provide answer-relevant information beyond generic length or stylistic effects. Here, we propose a trajectory-probing protocol for evaluating how a reasoning model's predicted answer evolves along a partial reasoning trace. The protocol (1) generates a model's full reasoning trace, (2) truncates it at fixed token-percentiles, and (3) injects each partial trace back into the model, measuring the model's induced answer distribution. We apply the protocol to seven open-source reasoning models (Qwen3-4B/8B/14B, gpt-oss-20b/120b, DeepSeek-R1-Distill-Llama-8B and Gemma-4-31B-it) across four benchmarks (GPQA Diamond, MMLU-Pro, Omni-MATH-2 and CompMath). The protocol reveals that accuracy and decision commitment generally increase as the percentage of provided reasoning tokens grows, and length, form, and token-identity controls confirm that gains stem from instance-specific semantic content rather than context length or generic reasoning style effects. Furthermore, cross-model probing shows that even stronger models frequently repeat the source model’s errors when given its partial reasoning trace, while allowing further reasoning substantially improves recovery. Together, these results provide a reproducible, model-agnostic measurement of reasoning trace dynamics that generalizes across answer formats and can reveal model-specific failure modes invisible to aggregate accuracy.
\end{abstract}

\begin{figure}[t]
	\centering
	\includegraphics[width=1\textwidth]{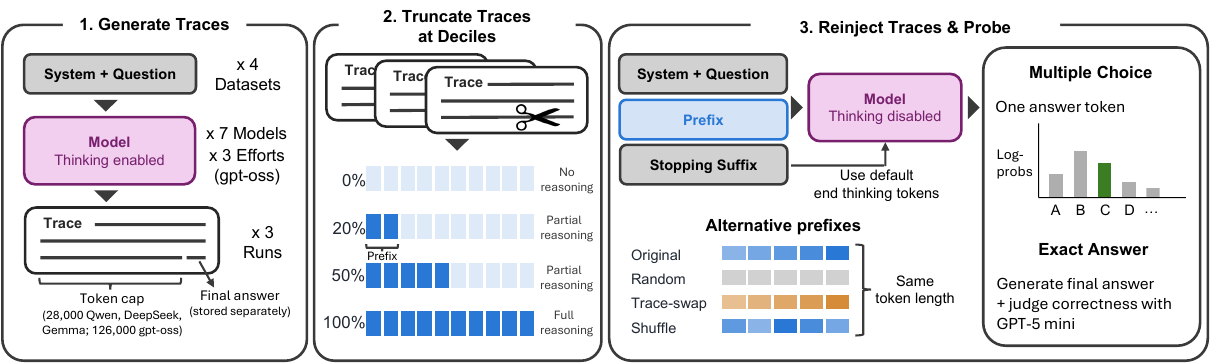}
\caption{Overview of the trajectory-probing protocol. (1) Generate full responses and extract reasoning traces. (2) Truncate each reasoning trace at fixed token deciles, including an empty-prefix baseline and the complete trace. (3) Reinject each prefix with the original question and an early-stopping suffix, then measure final answer probability and correctness. For exact-answer tasks, answer correctness is judged by GPT-5 mini and we average the answer's token probabilities. Length-matched random, swapped, and shuffled prefixes provide control conditions. The main evaluation covers seven models, four benchmarks, and three runs; reasoning effort ablations are restricted to the gpt-oss models on GPQA Diamond and MMLU-Pro.}
\label{fig:main_1}
\end{figure}

\section{Introduction}

The most capable large language models (LLMs) today are fine-tuned with reinforcement learning to produce a reasoning trace before delivering a final response ~\citep{agarwal2025gpt,jaech2024openai,yang2025qwen3}. The fine-tuning rewards are a combination of how well the LLM respects the output format (e.g., emitting <think> and </think> tags to encapsulate the reasoning trace in the Qwen3 model family~\cite{yang2025qwen3}) and the correctness of its final response~\citep{guan2024deliberative,guo2025deepseek,lambert2024tulu,wen2025reinforcement,wang2025beyond}. This approach builds on the Chain-of-Thought (CoT) methods~\cite{kojima2022zeroshotcot,wang2022selfconsistency,wei2022cot,yao2022react,yao2023tree,zelikman2022star}, where LLMs are encouraged to increase the amount of generated tokens, through prompting, tools, or chaining intermediate responses, before giving a final response. While the performance gains of scaling test-time compute are well established~\cite{muennighoff2025s1simpletesttimescaling,snell2024scaling}, there are remaining questions around the decision dynamics, including the effects on a model's final response calibration~\cite{kadavath2022language,vanhoyweghen2025lexical,xie2024calibrating,yoon2025reasoningconfidence}, the faithfulness of its reasoning traces~\cite{baker2025monitoring,chen2025reasoning,guan2025monitoring,kambhampati2025stop,korbak2025chain,lanham2023faithfulness,lyu2023faithful,stechly2025beyond,turpin2023unfaithful}, and the optimal amount of reasoning tokens~\cite{aytes2025sketch,ballon2025relationship,fan2025cothink,fu2025deep,liu-wang-2025-answer,sun2025stop}.

In this paper, we propose a trajectory-probing protocol for evaluating how a model's predicted answer evolves as progressively longer prefixes of a reasoning trace are revealed. The protocol turns a single reasoning trace into a sequence of partial-trace interventions, tracking accuracy and decision commitment (probability assigned to the model's eventual final answer) across reasoning depth. We compare the original prefixes with length-, form-, and token-identity-matched controls to assess the contribution of instance-specific reasoning content. Cross-model probing provides a complementary test: when a prefix leads its source model to an incorrect answer, does another model repeat that error, recover immediately, or recover after further reasoning? We also examine how the observed trajectories vary across difficulty strata, subject domains, and reasoning-effort settings, and assess their sensitivity to early-stopping suffix wording, deletion of answer expressions, and truncation at paragraph boundaries.

Prior work has substantially advanced our understanding of reasoning traces, including accuracy gains from scaling test-time compute and token-efficient reasoning~\cite{snell2024scaling,muennighoff2025s1simpletesttimescaling,aytes2025sketch,fan2025cothink,wang2025efficient}, intermediate stopping/commitment signals and step-importance analyses~\cite{liu-wang-2025-answer,sun2025stop,fu2025deep,bogdan2025thought}, and empirical evaluations of trace faithfulness and monitorability~\cite{turpin2023unfaithful,lanham2023faithfulness,chen2025reasoning,baker2025monitoring}. Related studies have also examined distinctions between the content and structural form of reasoning traces, showing benefits even when semantic coherence with the original instance is only partial~\cite{kambhampati2025stop,lanham2023faithfulness,stechly2025beyond}, and have investigated the reuse of reasoning traces across different models, for example, assessing whether weaker models benefit from reasoning produced by stronger models \citep{wang2025efficient,liu2025thought,lu2025reasoning,yang2025well,zelikman2022star}. Building upon truncation-based analyses of reasoning traces \citep{lanham2023faithfulness} and studies exploiting intermediate reasoning segments \citep{hammoud2025answerreasoningtraceuncovers}, we systematize these ideas into a trajectory-probing framework, which simultaneously tracks accuracy growth, decision commitment, and informational contribution across reasoning percentiles.

We demonstrate the reasoning-trace probing protocol on the multiple-choice benchmarks GPQA Diamond~\cite{rein2023gpqa}, MMLU-Pro~\cite{wang2024mmlupro} and the Vector Calculus subset of CompMath-MCQ~\cite{raimondi2026compmath}, and on the exact-answer benchmark Omni-MATH-2 \cite{ballon2026benchmarks}, with the open-source Qwen3-4B/8B/14B~\cite{yang2025qwen3}, gpt-oss-20b/120b~\cite{agarwal2025gpt}, DeepSeek-R1-Distill-Llama-8B~\cite{deepseekai2025deepseekr1} and Gemma-4-31B-it~\cite{gemmateam2026gemma4} models. We find that accuracy generally increases along the reasoning trajectory, consistent with the benefits of test-time compute and longer deliberation reported in the CoT and test-time scaling literature \citep{wei2022cot,snell2024scaling,muennighoff2025s1simpletesttimescaling}. At the same time, we observe non-trivial ``lost'' trajectories (right$\rightarrow$wrong) and a negative correlation between full-trace length and accuracy, even when accounting for the difficulty of the questions. Both observations connect to the current underthinking/overthinking debate \citep{chen2024not,wang2025thoughts,su2025between,ballon2025relationship}. Moreover, prediction confidence, defined as the probability assigned to the current highest-probability answer choice, generally rises for both correct and incorrect predictions. However, the mean confidence gap between correct and incorrect predictions does not improve with reasoning depth (and decreases on MMLU-Pro), indicating that longer traces can yield highly confident wrong answers~\cite{phan2025humanity,lyu2025calibrating,marjanovic2025deepseek,wei2024measuring}. Our trajectory probing also reveals model-specific anomalies: Qwen3-8B frequently generates \texttt{\textbackslash boxed\{\}} formatting, which biases the model toward predicting answer ``A'' and causes accuracy to drop from decile 90 to 100, indicating an idiosyncratic failure mode that would be invisible without systematic per-decile analysis.

Using length-, form-, and token-identity–matched controls, we further show that most of the accuracy gain is attributable to instance-specific semantic signal in the trace, with only a modest residual benefit surviving token permutation. These findings reconcile claims about ``reasonless'' intermediate tokens with evidence that coherent, instance-aligned reasoning matters \citep{jiang2025makes,kambhampati2025stop,lanham2023faithfulness,stechly2025beyond}. 
As a complementary test, cross-model probing examines prefixes that concern the same question but yield an incorrect source-model answer. Even stronger target models frequently repeat the source model's errors, while allowing further reasoning substantially improves recovery. A separate textual pilot finds that explicit backtracking is more common than restarting among correct departures (\Cref{tab:backtracking-pilot}), providing additional evidence that targets engage with the injected reasoning. These observations connect to work on thought reuse and recovery from unhelpful reasoning \citep{liu2025thought,lu2025reasoning,yang2025well,wang2025efficient}.

Together, these results show that the trajectory-probing protocol provides a reproducible, model-agnostic measurement of reasoning-trace dynamics that generalizes across multiple-choice and exact-answer benchmarks. The measurements are robust within difficulty strata, across early-stopping suffix wordings, and across reasoning-effort settings, and they surface model-specific failure modes that aggregate accuracy evaluations miss.

\begin{figure}[t]
    \centering
    \includegraphics[width=1\textwidth]{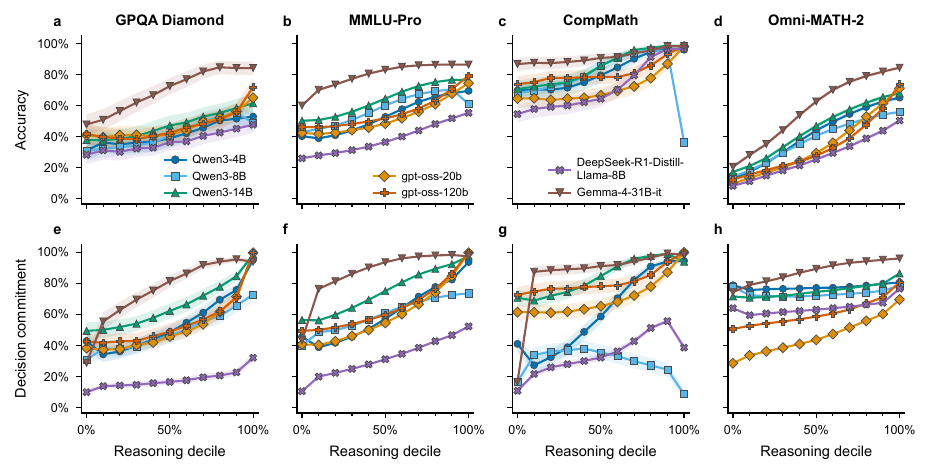}
 \caption{Accuracy and decision commitment across reasoning depth. \textbf{a--d}, Accuracy on GPQA Diamond ($n=198$), MMLU-Pro ($n=12{,}032$), CompMath ($n=267$), and Omni-MATH-2 ($n=4{,}181$), respectively. \textbf{e--h}, Corresponding decision-commitment. Curves show Qwen3-4B/8B/14B, gpt-oss-20b/120b, DeepSeek-R1-Distill-Llama-8B, and Gemma-4-31B-it, averaged over three runs. For multiple-choice tasks, accuracy uses the argmax over answer-letter probabilities and decision commitment is the unconditional probability assigned to the decile-100 predicted answer. For Omni-MATH-2, accuracy uses GPT-5 mini verdicts and decision commitment is the mean of generated-answer token probabilities. Shading shows pointwise 95\% question-bootstrap confidence intervals with 2{,}000 resamples, retaining all observed runs together. Accuracy generally increases, with a pronounced Qwen3-8B endpoint decline, especially on CompMath (\Cref{tab:boxed-collapse-full}). Companion non-choice and flip-rate panels for the multiple-choice settings are shown in \Cref{fig:appendix_main2_extra}.}
    \label{fig:main_2}
\end{figure}
\section{Probing the trajectories of reasoning traces}

We consider datasets with entries $(x_i,a_i)$, where $x_i$ is a question and $a_i$ is its reference answer. The reference answer $a_i$ is either a letter from a finite choice set $\mathcal{Y}$ for multiple-choice tasks, or a free-form string for exact-answer tasks. We evaluate GPQA Diamond ($n{=}198$, $\mathcal{Y}{=}\{A,B,C,D\}$), MMLU-Pro ($n{=}12{,}032$, $\mathcal{Y}{=}\{A,\ldots,J\}$), the Vector Calculus subset of CompMath-MCQ ($n{=}267$, $\mathcal{Y}{=}\{A,B,C\}$), and Omni-MATH-2 ($n{=}4{,}181$). Our seven models are Qwen3-4B/8B/14B, gpt-oss-20b/120b, DeepSeek-R1-Distill-Llama-8B, and Gemma-4-31B-it. All experiments use three independent runs (see \Cref{tab:appendix_variance} for variance across runs), unless explicitly indicated otherwise. See \Cref{sec:Appendix} for technical details on the experiments.

\subsection{The general protocol}
\label{subsec:protocol}
For each model $m$ and data entry $i$, we run a three-stage process as described in \Cref{fig:main_1}: 
\paragraph{Generate a full reasoning trace.} We prompt the model to solve the question with reasoning enabled, requesting a final response in the benchmark's answer format: a single choice letter for multiple-choice tasks or an answer string for exact-answer tasks. We then parse the generation to extract the reasoning trace $r_{m,i}$, separately from the final response (see \Cref{sec:Appendix} and the examples in \Cref{fig:example_gpt_full,fig:example_qwen_full}).
\paragraph{Slice the trace into token-based deciles.} Let $T_m(\cdot)$ be the tokenizer of model $m$, and let $L_{m,i}$ be the token length of its reasoning trace $r_{m,i}$. For $d\in\{10,20,\ldots,90\}$, we define $r^{(d)}_{m,i}$ by decoding the first $\lceil(d/100)L_{m,i}\rceil$ tokens. We set $r^{(0)}_{m,i}=\emptyset$ for the immediate-answer baseline and $r^{(100)}_{m,i}=r_{m,i}$ for the full-trace probe. Cross-model experiments use prefixes at $d\in\{20,40,60,80\}$, defined using the source model's tokenizer and transferred as text to the target model (see \Cref{subsec:rescuing protocol}).
\paragraph{Probe the induced answer distribution after injection} For each decile $d$, we construct a probe prompt that includes the system instruction, the original question $x_i$, the injected reasoning prefix $r^{(d)}_{m,i}$, and an early-stopping suffix that forces the model to stop reasoning (see \Cref{fig:fig:example_gpt_10,fig:example_qwen_10}). We then probe the answer distribution as described in \Cref{subsec:mc-probing} (multiple-choice) or \Cref{subsec:openended-probing} (exact-answer).

\subsection{Probing for multiple-choice answers}
\label{subsec:mc-probing}
For multiple-choice benchmarks, we generate one token after the probe prompt and obtain the next-token probabilities of the answer-choice letters. These probabilities are not renormalized over $\mathcal{Y}$, so probability mass assigned to other tokens remains visible. For each answer choice $y\in\mathcal{Y}$, we record
\(
p_{m,i,d}(y) = p_m\!\left(y \mid x_i, r^{(d)}_{m,i}\right).
\)
We take $\hat{y}^{(d)}_{m,i}=\arg\max_{y\in\mathcal{Y}}p_{m,i,d}(y)$ as the predicted answer at decile $d$. Accuracy measures whether this prediction equals $a_i$. Decision commitment is $p_{m,i,d}(\hat{y}^{(100)}_{m,i})$, the probability assigned to the choice predicted by the full-trace probe; it is an observable behavioral measure, not a claim about the model's internal state. We separately define prediction confidence as $p_{m,i,d}(\hat{y}^{(d)}_{m,i})$, the probability assigned to the current highest-probability choice. We compare its mean between correct and incorrect predictions in \Cref{fig:appendix_confidence}. We also measure non-choice probability, $1-\sum_{y\in\mathcal{Y}}p_{m,i,d}(y)$, and the flip rate between consecutive deciles. Comparing decile-0 and decile-100 predictions partitions trajectories into stable correct, gained (incorrect$\rightarrow$correct), lost (correct$\rightarrow$incorrect), and stable wrong.

\subsection{Probing for exact-answer benchmarks}
\label{subsec:openended-probing}
The same three-stage protocol applies to exact-answer benchmarks, where $a_i$ is a free-form string. Steps 1 and 2 are unchanged. In step 3, after constructing the probe prompt with the early-stopping suffix, we generate up to 100 answer tokens and obtain the per-token logprobabilities of the generated sequence. We append a short instruction-following suffix (``My final answer is:'') after the early-stopping marker to prompt the model toward emitting an answer rather than continuing reasoning. We use GPT-5 mini to assess equivalence between the model's final answer and the reference answer, with prompting conventions as in \citep{ballon2026benchmarks}. To measure decision commitment, we compute the average log-probabilities of the model's final answer.

\subsection{Controls for context length, trace structure, and token identity}
\label{subsec:controls}
Injected prefixes can affect answer probabilities through their length, generic reasoning style, or individual tokens, in addition to content relevant to the current question. To assess these contributions, we compare each original prefix with three controls matched in token length at each intervention point. These experiments cover all seven models on GPQA Diamond, MMLU-Pro, and CompMath, with three runs per condition (\Cref{sec:Appendix}):

\begin{itemize}
\item Random control: Replace $r^{(d)}_{m,i}$ with an equally long sequence of randomly sampled tokens under $T_m$, testing whether an arbitrary prefix of the same length produces comparable effects.
\item Swap control: Replace $r^{(d)}_{m,i}$ with a reasoning prefix from a different question, truncated to the same token length under $T_m$. This retains the form of a real reasoning trace while breaking its alignment with the current question.
\item Shuffle control: Replace $r^{(d)}_{m,i}$ with a random permutation of its tokens, preserving their identities and count while disrupting their sequential organization.
\end{itemize}

We again probe the induced answer distributions as described in step 3 of the probing protocol (see \Cref{subsec:protocol}).

\subsection{Cross-model probing of prefixes yielding incorrect answers}
\label{subsec:rescuing protocol}

The probing protocol set out in \Cref{subsec:protocol} also supports cross-model experiments.

\begin{wrapfigure}[25]{r}{54mm}
\vspace{-4mm}
    \centering
\includegraphics[width=46mm]{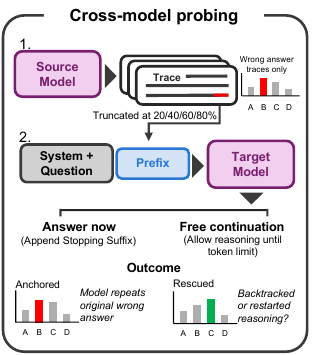}
\caption{Cross-model probing of incorrect partial traces. Select source-model prefixes yielding an incorrect answer at 20\%, 40\%, 60\%, or 80\% of the reasoning trace, then inject them into a target model. Compare an immediate answer with free continuation. Rescue denotes a correct target answer; anchoring denotes repetition of the source model's incorrect answer. The target may also produce a different incorrect answer.}
    \label{fig:cross_model_probing}
\end{wrapfigure}
Fix a source model $m_b$ and a target model $m_t$. We evaluate ten forward pairs among Qwen3-4B/8B/14B and gpt-oss-20b/120b, together with two reverse directions: Qwen3-8B$\to$Qwen3-4B and gpt-oss-20b$\to$Qwen3-14B. These reverse transfers account for accuracy rankings that do not consistently follow model size: Qwen3-4B exceeds Qwen3-8B at full trace on both evaluated benchmarks, while Qwen3-14B exceeds gpt-oss-20b on MMLU-Pro (\Cref{fig:main_2}). Transfers are evaluated on GPQA Diamond using runs 1--3 and on MMLU-Pro using run 1, at prefix deciles $d\in\{20,40,60,80\}$. For each question $x_i$ and decile $d$, let $\hat{y}^{(d)}_{b,i}=\arg\max_y p_{b,i,d}(y)$ be the initial model's final response.
We only take into account questions for which $\hat{y}^{(d)}_{b,i}\neq a_i$, i.e. incorrect responses at decile $d$. For each such sample, we inject the corresponding slice $r^{(d)}_{b,i}$ into the target model's prompt and evaluate two conditions (see \Cref{sec:Appendix}): 

In the answer now condition, we inject $r^{(d)}_{b,i}$ into the target model and append the appropriate early-stopping suffix to obtain an immediate answer distribution $p^{\text{now}}_{t,i,d}$ and prediction $\hat{y}^{\text{now}}_{t,i,d}$. In the free continuation condition, we allow the target to extend the injected prefix subject to context limits before appending the early-stopping suffix and probing its answer distribution. This yields $p^{\text{free}}_{t,i,d}$ and prediction $\hat{y}^{\text{free}}_{t,i,d}$.

This comparison measures immediate recovery versus recovery after additional reasoning from the same source-error prefix. It provides a complementary control using reasoning about the same question that yields an incorrect source prediction. Successful recovery alone does not distinguish backtracking from restarting or following another route; a separate textual pilot examines these behaviors (\Cref{tab:backtracking-pilot}). We define the rescue rate as the probability that the target answers correctly, conditional on an incorrect source prediction:
\[
\mathrm{Rescue}(b \to t, d) = P\bigl(\hat{y}_{t,i,d} = a_i \;\big|\; \hat{y}_{b,i,d} \neq a_i\bigr),
\]
and the anchoring rate as the probability that the target repeats the initial model's wrong answer:
\[
\mathrm{Anchor}(b \to t, d) = P\bigl(\hat{y}_{t,i,d} = \hat{y}_{b,i,d} \;\big|\; \hat{y}_{b,i,d} \neq a_i\bigr),
\]
where $\hat{y}_{m,i,d} = \arg\max_y p_{m,i,d}(y)$ is the argmax prediction. Both metrics are computed separately for base and free modes. Note that these probabilities do not sum to one, as the target model may pick another incorrect response. 

\begin{figure}[h!]
	\centering
	\includegraphics[width=1\textwidth]{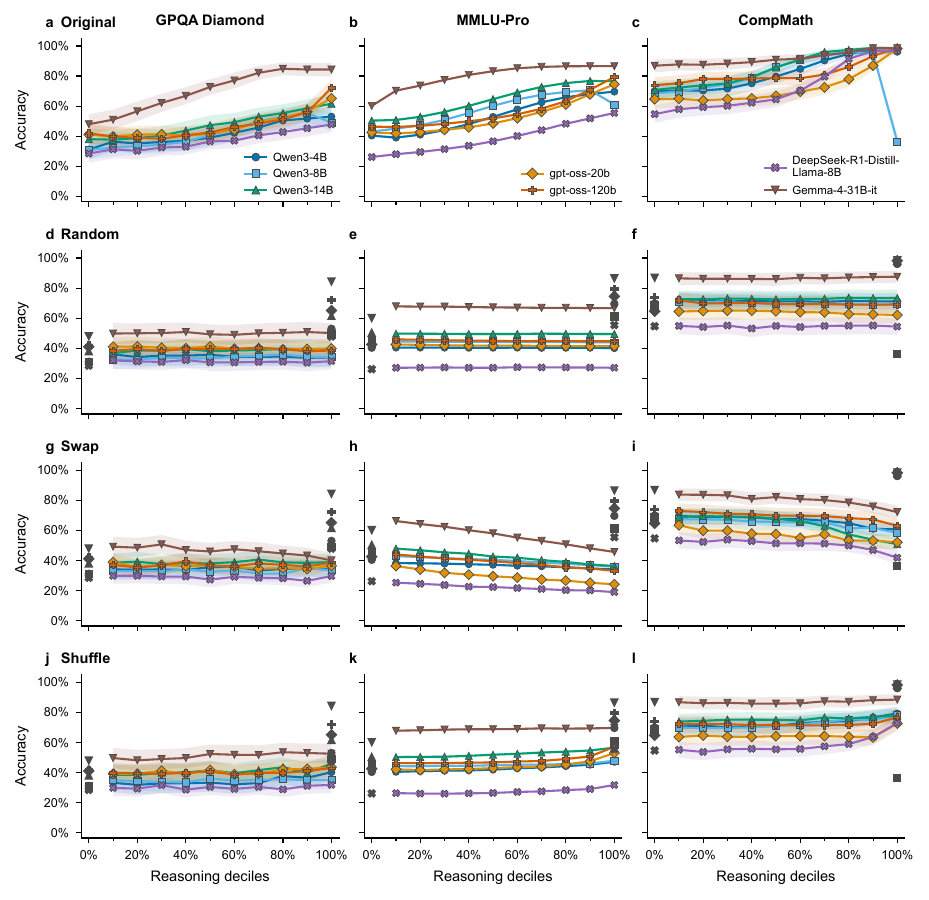}
\caption{Original reasoning traces compared with length-matched controls. Columns show GPQA Diamond, MMLU-Pro, and CompMath; rows show original prefixes (\textbf{a--c}), random tokens (\textbf{d--f}), another question's reasoning prefix (\textbf{g--i}), and a permutation of the original prefix's tokens (\textbf{j--l}). Curves show the seven models in \Cref{fig:main_2}, averaged over three runs. Endpoint markers in the control panels indicate the original no-reasoning and full-trace accuracies. Shading shows pointwise 95\% question-bootstrap confidence intervals with 2{,}000 resamples, retaining paired runs and conditions together. Original traces generally outperform the controls at full length; Qwen3-8B on CompMath is an exception, where its endpoint collapse puts original accuracy below all three controls. Paired endpoint comparisons and their statistical conventions are reported in \Cref{tab:summary-statistics}.}
	\label{fig:main_3}
\end{figure}

\section{Results}
\subsection{Accuracy and decision commitment increase overall with reasoning depth}

Accuracy generally increases with the proportion of reasoning tokens provided, with higher full-trace than no-reasoning accuracy in 27 of the 28 model--benchmark combinations (\Cref{fig:main_2}a--d). On GPQA Diamond, models improve from 28.5--48.0\% at decile 0 to 47.8--84.3\% at decile 100. On MMLU-Pro, from 26.2--60.2\% to 55.5--86.6\% (see \Cref{tab:summary-statistics} for full statistics). On CompMath, accuracy ranges from 54.7--86.9\% at decile 0 to 36.5--98.6\% at decile 100. On Omni-MATH-2, judge-scored correctness rises from 8.4--20.7\% at decile 0 to 50.5--84.6\% at decile 100. The magnitude and timing of accuracy gains vary across models and benchmarks. The notable exception is Qwen3-8B, which shows a substantial decline going from decile 90 to 100; on CompMath, accuracy drops from 95.3\% to 36.5\%. This anomaly stems from Qwen3-8B frequently generating \texttt{\textbackslash boxed\{\}} formatting (27\% of traces), which biases the model toward predicting answer ``A'' and substantially reduces accuracy (see \Cref{fig:appendix_non_choice} and \Cref{tab:boxed-collapse-full}). Decision commitment follows a similar trajectory: the probability assigned to the model's eventual final response (i.e. at decile $100\%$) rises overall (\Cref{fig:main_2}e--h), while non-choice probability (mass on non-answer tokens) shows model- and benchmark-specific trends (\Cref{fig:appendix_main2_extra}a--c). On CompMath, non-choice probability remains near zero for the gpt-oss models, but increases from 33.3\% to 91.5\% for Qwen3-8B and from 43.3\% to 61.0\% for DeepSeek-R1-Distill-Llama-8B between deciles 90 and 100 (\Cref{fig:appendix_main2_extra}c). The flip rate (how often the argmax prediction changes between consecutive deciles) generally decreases with reasoning depth for MMLU-Pro and remains mostly stable (except for deciles 10 and 100) in GPQA Diamond (\Cref{fig:appendix_main2_extra}d,e). On CompMath, flip rates show model-specific patterns rather than a consistent decline: the gpt-oss models show higher rates in later portions of the trace, while Qwen3-8B exhibits a sharp endpoint spike, with 61.7\% of predictions changing between deciles 90 and 100 (\Cref{fig:appendix_main2_extra}f). These patterns are robust to alternative early-stopping prompts and across reasoning efforts (see \Cref{fig:appendix_suffix,fig:appendix_effort,fig:appendix_effort_trajectories}).  Additional answer-deletion and paragraph-boundary experiments on GPQA Diamond and MMLU-Pro preserve broad trajectory patterns while revealing model-specific sensitivities (\Cref{sec:Appendix}).

Comparing decile-0 and decile-100 predictions, gained trajectories (incorrect$\rightarrow$correct) outnumber lost trajectories (correct$\rightarrow$incorrect) for all model--benchmark combinations except Qwen3-8B on CompMath. Across models, lost trajectories account for 3.5--11.3\% of questions on GPQA Diamond and 2.2--6.3\% on MMLU-Pro. On CompMath, loss rates are 0.1--2.0\% for the other six models, whereas Qwen3-8B has 35.2\% lost and 3.1\% gained trajectories, reflecting its full-trace accuracy collapse (\Cref{fig:appendix_trajectories}). On Omni-MATH-2, loss rates are 1.1--2.1\%, compared with gained fractions of 43.5--65.0\%; outcomes within difficulty strata are shown in \Cref{fig:appendix_omnimath_trajectories}. Prediction confidence, defined as the probability assigned to the current highest-probability answer choice, often rises for both correct and incorrect multiple-choice predictions. The discrimination gap (mean prediction confidence when correct minus when incorrect) varies across models on GPQA Diamond and CompMath, and narrows for most models on MMLU-Pro. Thus, additional reasoning does not consistently improve confidence discrimination and can yield incorrect answers assigned high probability (\Cref{fig:appendix_confidence}).

Shorter reasoning traces are generally associated with higher accuracy on GPQA Diamond and MMLU-Pro (\Cref{fig:appendix_accuracy_length}d,e), suggesting that models produce extended reasoning when struggling with a question. On CompMath, all models except Qwen3-8B maintain near-ceiling accuracy across the first four trace-length quintiles, with lower accuracy in the longest quintile; Qwen3-8B instead shows a U-shaped relationship (\Cref{fig:appendix_accuracy_length}f). On Omni-MATH-2, the overall negative association between trace length and accuracy persists within difficulty strata for all six other models, while Qwen3-8B again shows a non-monotonic pattern (\Cref{fig:appendix_omnimath_accuracy_length}). On MMLU-Pro, we also see substantial variation in reasoning dynamics across domains in the dataset (\Cref{fig:appendix_stratified_mmlu_categories}).


\subsection{Instance-specific semantic reasoning drives accuracy gains}
To assess which aspects of reasoning traces contribute to accuracy gains, we compare original prefixes with three length-matched controls across all seven models on GPQA Diamond, MMLU-Pro, and CompMath (\Cref{fig:main_3}). At full trace, original accuracy exceeds all three controls in 20 of the 21 model--benchmark combinations; the exception is Qwen3-8B on CompMath, where the endpoint collapse places original accuracy below every control. Excluding this exception, original traces improve accuracy by 11.6--42.1 percentage points relative to the no-reasoning baseline. Across all model--benchmark combinations, random controls change accuracy by $-4.7$ to $+6.5$ percentage points, indicating that context length alone does not reproduce the gains from original reasoning. Swap controls change accuracy by $-20.0$ to $+2.9$ percentage points, showing that a real reasoning trace from another question generally provides little benefit or reduces accuracy. Shuffle controls yield gains of 1.0--18.2 percentage points, suggesting that individual token content retains useful information even after sequential order is disrupted, while the larger gains from original traces demonstrate the additional contribution of coherent, instance-aligned reasoning. Paired full-trace comparisons between original and control conditions are reported in \Cref{tab:summary-statistics} (all $p<0.001$, with the direction reversed for Qwen3-8B on CompMath); these McNemar tests pool question--run observations without adjusting for dependence across repeated runs of the same question.


Overall, these controls support the conclusion that reasoning traces carry instance-specific semantic information that improves answer accuracy beyond the effects of context length, generic reasoning style, or token identity alone. 

\begin{figure}[h!]
	\centering
	\includegraphics[width=1\textwidth]{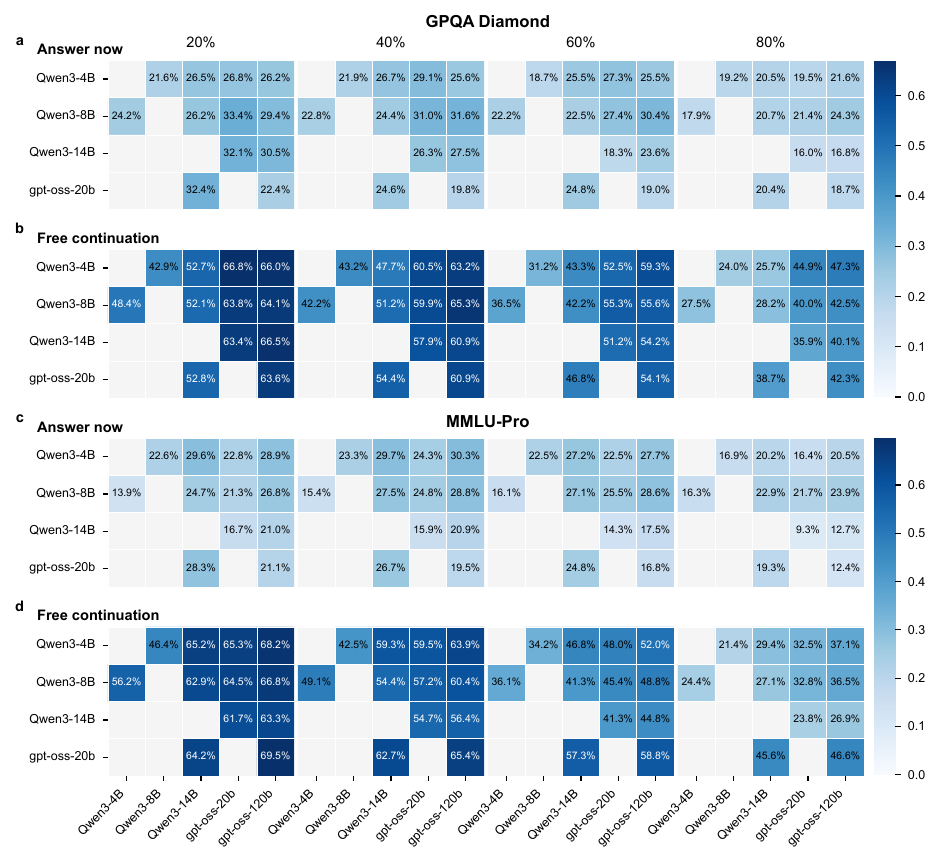}
\caption{Rescue rates under immediate answering and free continuation. Each cell reports the percentage of selected transfers for which the target answers correctly, conditional on the source model answering incorrectly at that prefix. Columns of heatmaps correspond to 20\%, 40\%, 60\%, and 80\% prefixes; within each heatmap, rows identify source models and columns identify targets. \textbf{a,b}, GPQA Diamond, pooling measured transfers from runs 1--3. \textbf{c,d}, MMLU-Pro, run 1. Panels \textbf{a,c} use answer now; \textbf{b,d} allow free continuation. The figure includes ten forward pairs and two reverse pairs per benchmark (Qwen3-8B$\to$Qwen3-4B and gpt-oss-20b$\to$Qwen3-14B). These reverse directions account for accuracy rankings that do not consistently follow model size: Qwen3-4B exceeds Qwen3-8B at full trace on both benchmarks, while Qwen3-14B exceeds gpt-oss-20b on MMLU-Pro (\Cref{fig:main_2}). Blank cells denote unmeasured pairs. Two MMLU-Pro reverse-transfer measurements exceed the target context limit and are omitted. Free continuation improves recovery, with lower rescue rates at later prefixes for many pairs. For a given source model and prefix, gpt-oss targets show higher observed rescue rates than the evaluated Qwen3 targets under free continuation; this family ordering is less consistent when answering immediately. \Cref{tab:rescue-summary} summarizes the ten forward pairs only. In a separate continuation-trace pilot, explicit backtracking was more common than restarting among correct departures (\Cref{tab:backtracking-pilot}).}
	\label{fig:main_4}
\end{figure}

\subsection{Stronger models can rescue incorrect reasoning, and free continuation substantially improves recovery}

Cross-model probing provides a complementary control using same-question prefixes that yield an incorrect source-model answer. Across the twelve evaluated transfer pairs, free continuation improves recovery compared with answering immediately (\Cref{fig:main_4}). On GPQA Diamond, pooling runs 1--3, answer-now rescue rates range from 21.6--33.4\% at decile 20 to 16.0--24.3\% at decile 80; allowing further reasoning raises these ranges to 42.9--66.8\% and 24.0--47.3\%, respectively. On MMLU-Pro, evaluated on run 1, the corresponding answer-now ranges are 13.9--29.6\% and 9.3--23.9\%, compared with 46.4--69.5\% and 21.4--46.6\% under free continuation. Rescue rates are generally lower for later prefixes. In matched comparisons using the same source model and prefix decile, gpt-oss targets achieve higher observed rescue rates than the evaluated Qwen3 targets under free continuation; this ordering is less consistent when answering immediately. For the ten forward pairs summarized in \Cref{tab:rescue-summary}, pooling observations across the four evaluated deciles gives rescue gains of 15.7--35.8 percentage points on GPQA Diamond and 16.3--43.4 percentage points on MMLU-Pro (all $p<0.001$, paired McNemar tests). These pooled tests do not adjust for repeated observations of a question across deciles or runs.

Despite successful rescues, target models frequently repeat the source model's incorrect answer, and anchoring is greater at later prefixes (\Cref{fig:appendix_anchoring}). Across all twelve pairs on GPQA Diamond, answer-now anchoring ranges from 26.2--49.1\% at decile 20 to 50.7--66.4\% at decile 80; with free continuation, these ranges are 15.3--32.2\% and 31.2--55.8\%. On MMLU-Pro, answer-now anchoring ranges from 28.5--42.1\% at decile 20 to 56.1--74.8\% at decile 80, compared with 13.0--23.7\% and 40.0--60.3\% under free continuation. For the ten forward pairs in \Cref{tab:rescue-summary}, free continuation reduces pooled anchoring by 9.8--31.4 percentage points on GPQA Diamond and 11.3--26.7 percentage points on MMLU-Pro (all $p<0.001$, under the same pooled-testing convention). Thus, further reasoning improves recovery and reduces anchoring, but even stronger targets frequently repeat source-model errors. Because transfers are selected on source errors, rescue rates alone do not establish a performance loss relative to the target's no-prefix accuracy on the same selected questions.

To examine how recovery occurs, a separate textual pilot analyzes 9{,}529 continuations, including 4{,}921 correct continuations and 1{,}079 correct departures from the injected reasoning. Among these correct departures, 80.9\% are classified as explicit backtracking, 18.1\% as changing direction, and 1.0\% as restarting (\Cref{tab:backtracking-pilot}). These percentages describe the correct-departure subset, not all successful rescues. The predominance of explicit backtracking supports the interpretation that recovery can involve revising the injected reasoning, although textual behavior does not establish faithfulness to the model's internal computation.

\section{Discussion}

Our trajectory-probing protocol provides a reproducible, model-agnostic measurement of how predicted answers evolve along partial reasoning traces, across multiple-choice and exact-answer benchmarks. Across seven models and four benchmarks, accuracy generally increases with reasoning depth: full-trace accuracy exceeds the no-reasoning baseline in 27 of 28 model--benchmark combinations, with positive gains of 11.6--63.9 percentage points, consistent with the benefits of test-time computation~\citep{wei2022cot,snell2024scaling,muennighoff2025s1simpletesttimescaling}. Length-, form-, and token-identity-matched controls support a contribution from instance-specific semantic content beyond context length or generic reasoning style~\citep{kambhampati2025stop,lanham2023faithfulness,stechly2025beyond}. Cross-model probing provides a complementary test: even stronger targets frequently repeat errors elicited by same-question source prefixes, while free continuation improves rescue rates by 15.7--43.4 percentage points across the ten forward pairs and two benchmarks summarized in \Cref{tab:rescue-summary}. The separate textual pilot further shows that successful departures can involve explicit revision of the injected reasoning, connecting these findings to work on thought reuse and recovery~\citep{liu2025thought,lu2025reasoning,yang2025well,wang2025efficient}.

These average gains coexist with failures that aggregate accuracy can obscure. Lost trajectories account for 3.5--11.3\% of questions on GPQA Diamond, and increasing prediction confidence does not consistently improve discrimination between correct and incorrect answers. Revealing more of a given trace can therefore improve accuracy overall while still producing incorrect answers assigned high probability~\citep{phan2025humanity,lyu2025calibrating,marjanovic2025deepseek,wei2024measuring}. Across questions, longer reasoning traces are generally associated with lower accuracy, including within Omni-MATH-2 difficulty strata for the six models other than Qwen3-8B. This association is consistent with models reasoning longer when struggling, without implying that shortening those traces would improve their answers~\citep{chen2024not,wang2025thoughts,su2025between,ballon2025relationship}. Trajectory probing also localizes the Qwen3-8B failure mode: on CompMath, accuracy drops from 95.3\% to 36.5\% between deciles 90 and 100. For cross-model reuse, later prefixes generally permit less recovery and exhibit greater anchoring, even with further reasoning. These findings motivate evaluating when to stop reasoning~\citep{liu-wang-2025-answer,sun2025stop,fu2025deep}, monitoring intermediate predictions~\citep{baker2025monitoring,guan2025monitoring,korbak2025chain}, and allowing recovery opportunities when transferring traces between models~\citep{su2025between,stechly2025beyond,liu2025thought}.

\paragraph{Limitations}
Several limitations bound our conclusions. The seven evaluated models cover Qwen3, gpt-oss, DeepSeek-R1-Distill-Llama, and Gemma, but do not establish generality to closed-source frontier models. All main trajectory experiments use three runs; cross-model probing remains limited to the Qwen3 and gpt-oss models on GPQA Diamond and MMLU-Pro, with only one transfer run on MMLU-Pro. The length-matched controls cover multiple-choice tasks, while exact-answer accuracy on Omni-MATH-2 depends on GPT-5 mini judgments. Pooled McNemar tests do not adjust for repeated observations of the same question across runs or, for transfers, deciles. The backtracking pilot provides evidence about textual continuation behavior, but its small manual agreement check does not validate every classification or establish that these behaviors generalize to all rescues. Finally, the protocol measures the functional effects of trace injection on model outputs; neither these effects nor explicit textual backtracking establish faithfulness to the model's internal computations~\citep{turpin2023unfaithful,lanham2023faithfulness,chen2025reasoning}.

\section*{Acknowledgements}
This work received funding from the Flemish Government's ``Onderzoeksprogramma Artifici\"ele Intelligentie (AI) Vlaanderen'' program. Andres Algaba acknowledges support from the Vrije Universiteit Brussel Research Council (VUB-OZR), the Francqui Foundation (Belgium) through a Francqui Start-Up Grant, and a fellowship from the Research Foundation Flanders (FWO) under Grant No.1286924N. Vincent Ginis acknowledges support from Research Foundation Flanders under Grant No.G032822N and G0K9322N. Computing resources and services were supplied by the VSC (Flemish Supercomputer Center), with funding from the Research Foundation Flanders (FWO) and the Flemish Government.

\section*{Author contributions}
Andres Algaba, Marthe Ballon, and Vincent Ginis conceived the study. Andres Algaba and Marthe Ballon performed the analysis, prepared the figures, and wrote the initial manuscript. All authors contributed to manuscript revision and critical feedback.

\section*{Data and code availability}
Code is available at \url{https://github.com/AndresAlgaba/probing_reasoning_traces}. Data and outputs associated with this study are available at \url{https://doi.org/10.5281/zenodo.18430487}.

\clearpage
\bibliographystyle{unsrt}
\bibliography{references}


\clearpage
\appendix

\setcounter{figure}{0}
\renewcommand{\thefigure}{A\arabic{figure}}
\renewcommand{\theHfigure}{A.\arabic{figure}}

\setcounter{table}{0}
\renewcommand{\thetable}{A\arabic{table}}
\renewcommand{\theHtable}{A.\arabic{table}}

\section{Technical details}
\label{sec:Appendix}


\subsection{Datasets}

\paragraph{GPQA Diamond.} We use the GPQA Diamond dataset~\cite{rein2023gpqa}, a challenging benchmark of 198 graduate-level multiple-choice questions in biology, chemistry, and physics. Questions were written by domain experts and validated to be difficult even for experts outside their specialty, with human expert accuracy around 65\%. Each question has four answer choices (A--D). GPQA Diamond represents the highest-quality subset of the broader GPQA dataset, selected for question clarity and expert agreement.

\paragraph{MMLU-Pro.} We use the MMLU-Pro dataset~\cite{wang2024mmlupro}, an enhanced version of the Massive Multitask Language Understanding benchmark containing 12,032 questions across 14 subject categories spanning STEM (biology, chemistry, computer science, engineering, math, physics), humanities (history, law, philosophy), and social sciences (business, economics, health, psychology, other). Unlike the original MMLU with four answer choices, MMLU-Pro expands to ten choices (A--J) per question, reducing the effectiveness of random guessing and increasing difficulty. Questions were filtered to remove trivial items and augmented with more challenging, reasoning-focused problems.

\paragraph{Omni-MATH-2} We use the Omni-MATH-2 dataset, an open-ended graduate-level mathematics problems containing $4181$ cleaned problems compared to the original Omni-MATH dataset~\citep{gao2024omni,ballon2025estimating}. The dataset spans over 33 subdomains and 10 difficulty levels (predefined difficulty ratings ranging from a 1--9.5). We divide the data into four difficulty tiers based on the percentiles of the difficulty distribution to create balanced classes. Answers are free-form expressions (numbers, equations, sets) rather than multiple-choice letters, and correctness is judged by GPT-5 mini (default) using the same judge prompt as in \cite{ballon2026benchmarks}.

\subsection{Models}
We evaluate five LLMs from two model families: Qwen3-4B, Qwen3-8B, and Qwen3-14B from the Qwen3 family~\citep{yang2025qwen3}, and gpt-oss-20b and gpt-oss-120b from the gpt-oss family~\citep{agarwal2025gpt}.

\paragraph{Model context lengths.} We configure each model with its maximum supported context length: 32,768 tokens for all Qwen3 models (native supported, not using YaRN scaling~\cite{peng2023yarn}) and 131,072 tokens for the gpt-oss models. These context lengths accommodate full reasoning traces plus the question prompt and system instructions.

\paragraph{Trace generation.} For initial reasoning trace generation (full thinking mode), we use \texttt{max\_tokens=28,000} for Qwen3 models and \texttt{max\_tokens=126,000} for gpt-oss models. Sampling parameters follow each model's recommended settings: Qwen3 models use \texttt{temperature=0.6}, \texttt{top\_p=0.95}, \texttt{top\_k=20}, and gpt-oss models use \texttt{temperature=1}, \texttt{top\_p=1}, \texttt{top\_k=50}. Generation terminates when either the \texttt{max\_tokens} limit is reached or the model emits an end-of-sequence token.

\begin{figure}[t]
	\centering
	\includegraphics[width=1\textwidth]{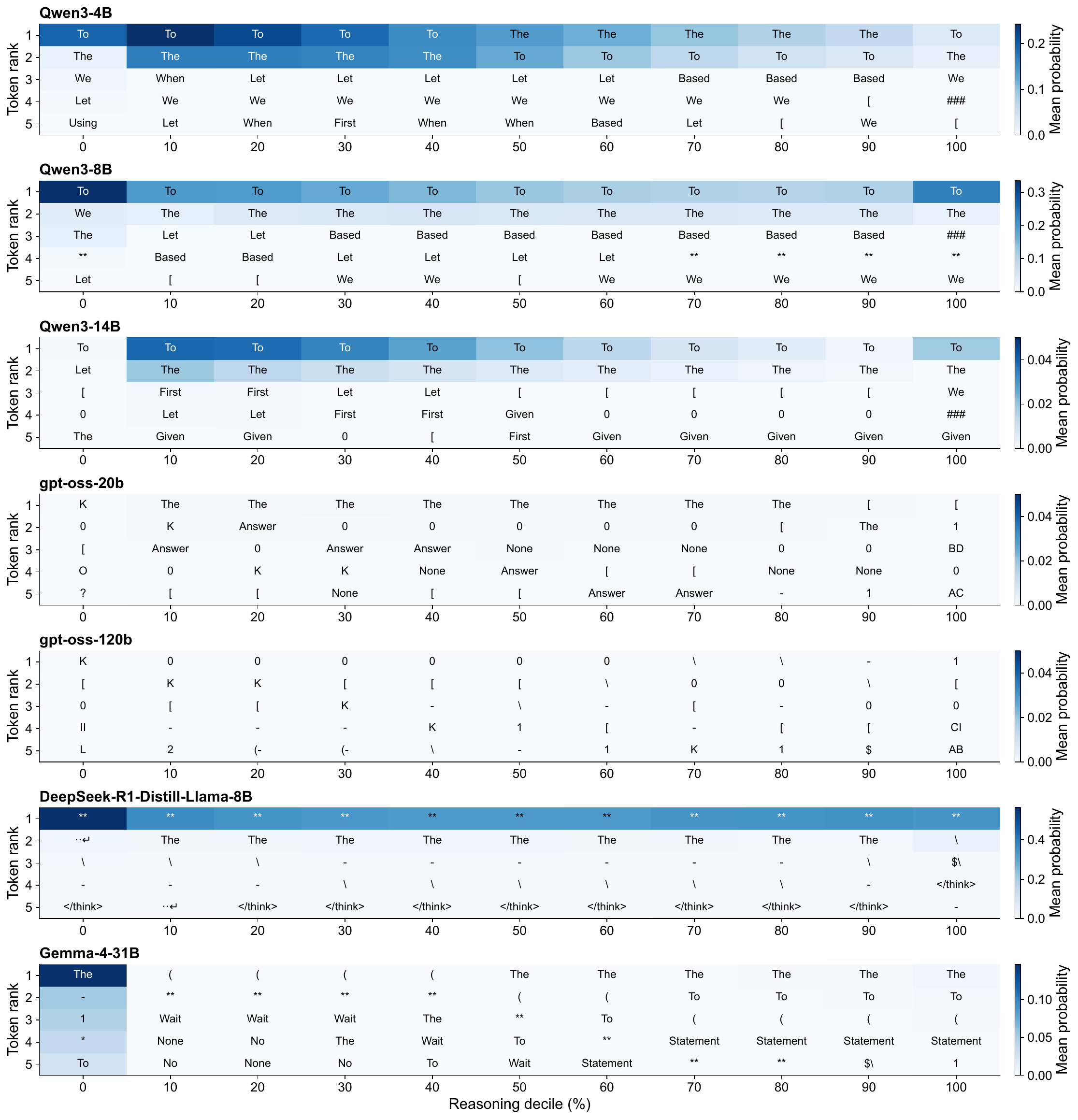}
\caption{Highest-probability non-choice tokens by model and decile. For each of the seven models, tokens are ranked by their mean recorded probability across GPQA Diamond and MMLU-Pro question--run observations from runs 1--3. Five tokens are displayed at each decile. Tokens absent from a returned probability record contribute zero. Cell shading indicates mean probability, with a separate color scale for each model. The pooled means weight question--run observations equally, so the larger MMLU-Pro dataset contributes more observations.}
	\label{fig:appendix_non_choice}
\end{figure}

\paragraph{Answer probing and probability extraction.} During the probing step (step 3 of the protocol in \Cref{subsec:protocol}), the reported probabilities are raw softmax outputs over logits, unaffected by temperature, top-$k$, or top-$p$ settings. We request the model's next-token log-probabilities at the answer position and convert to probabilities via softmax. No temperature scaling or nucleus/top-$k$ filtering is applied, preserving the model's true probability assignments over answer choices. We define the choice-token set as bare uppercase letters (``A'', ``B'', etc.) . We have also considered variants, such as with leading spaces (e.g., `` A''), but they all tokenize to multiple tokens in all tested models and rarely appear as the first generated token (see \Cref{fig:appendix_non_choice}).

\subsection{Prompts}
We use model-specific prompt formats to ensure compatibility with each model family's expected input structure.

\paragraph{System instruction.} All models receive the same core instruction requesting a single-letter answer:
\begin{quote}
\textit{Solve the following problem. Please make sure that your response only consists of a single letter corresponding to the correct answer choice. Do not include anything else in your final response.}
\end{quote}
For Omni-MATH-2 we use:
\begin{quote}
\textit{Solve the following problem. Please make sure that your response only consists of your final answer. Do not include anything else in your final response.}
\end{quote}

\paragraph{Prompt formatting.} The gpt-oss family~\citep{agarwal2025gpt} uses a structured format with system, developer, and user message tags, along with channel annotations (analysis, commentary, final) to separate reasoning from the final response (see \Cref{fig:example_gpt_format} for the full prompt structure and \Cref{fig:example_gpt_full} for an example complete response). For the no-reasoning baseline (decile 0), we force an immediate final response by closing the analysis channel before any reasoning occurs (\Cref{fig:example_gpt_no_reason_format}).

The Qwen3 family~\citep{yang2025qwen3} uses a simpler format with \texttt{<|im\_start|>} and \texttt{<|im\_end|>} tags, and encapsulates reasoning within \texttt{<think>} and \texttt{</think>} tags (see \Cref{fig:example_qwen_format} for the prompt structure and \Cref{fig:example_qwen_full} for an example complete response). For the no-reasoning baseline, we include empty thinking tags (\Cref{fig:example_qwen_no_reason_format}).

\begin{figure}[t]
    \centering
    \includegraphics[width=1\textwidth]{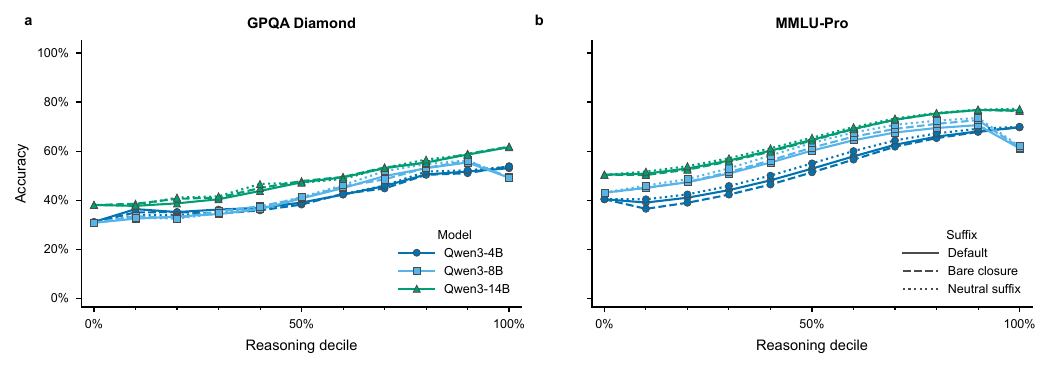}
    \caption{Early-stopping suffix ablation for the Qwen3 family. Per-decile accuracy on GPQA Diamond (left) and MMLU-Pro (right), averaged over 3 runs, with three suffix variants: default (Qwen documentation), minimal (\texttt{</think>} only), and neutral (``Based on my reasoning so far, my answer is:''). Trajectories are nearly indistinguishable across the three suffixes, confirming the protocol's measurements are not driven by suffix phrasing. Note that forcing reasoning to stop after a pre-specified number of tokens by appending a suffix is exactly how the Qwen3 documentation enforces this, and how it explicitly trains the model to handle these cases \citep{yang2025qwen3}. The gpt-oss family is excluded as it uses native API structural tokens with no linguistic suffix.}
    \label{fig:appendix_suffix}
\end{figure}

\paragraph{Early-stopping suffixes.} To probe the model's answer distribution at intermediate reasoning deciles, we inject a partial reasoning trace followed by an early-stopping suffix that forces the model to emit a final response. For the Qwen3 family, we use the suggested early-stopping suffix~\citep{yang2025qwen3}:
\begin{quote}
\textit{Considering the limited time by the user, I have to give the solution based on the thinking directly now.\textbackslash n</think>\textbackslash n\textbackslash n}
\end{quote}
For Omni-MATH-2, we append the following string after the chosen suffix "My final answer is:" to prompt the model to give a final answer rather than continuing with an explanation/justification.
We also evaluate alternative suffixes, one that simply closes the thinking mode (``\textbackslash n</think>\textbackslash n\textbackslash n''), and one that says (``Based on my reasoning so far, my answer is:'') which yield comparable results (see \Cref{fig:appendix_suffix}). \Cref{fig:example_qwen_10} shows an example of a 10\% reasoning decile probe with the full early-stopping suffix, and \Cref{fig:example_qwen_10_alternative} shows the same with the alternative suffix. For the gpt-oss family, we close the analysis channel and open the final channel to force an immediate answer (\Cref{fig:fig:example_gpt_10}).

\subsection{Controls}
In \Cref{subsec:controls}, we introduce controls to isolate the contribution of reasoning content from confounding factors such as token count, we implement three control conditions.

\paragraph{Random control.} We generate random token sequences matching the exact token length of the original reasoning trace. Tokens are sampled uniformly at random from the model's vocabulary, excluding special tokens (e.g., \texttt{<think>}, \texttt{</think>}, \texttt{<|im\_start|>}).

\paragraph{Swap control.} For each example $i$, we inject the reasoning trace from a different example $j \neq i$. The source example $j$ is sampled uniformly at random from examples whose reasoning trace has token length $\geq$ the target length, then truncated from the beginning to exactly match the target's token count (i.e., we take the first $k$ tokens where $k$ is the target prefix length). This ensures: (1) coherent, well-formed reasoning text, (2) reasoning for the wrong question, and (3) identical token count to control for length effects. This control tests whether coherent but irrelevant reasoning helps.

\paragraph{Shuffle control.} We randomly permute the tokens of the original reasoning trace, preserving the exact token multiset but destroying sequential coherence. The thinking-start marker (e.g., \texttt{<think>}) is excluded from shuffling to maintain valid prompt structure. This control tests whether the specific token ordering matters for performance, rather than just the presence of reasoning-related tokens.

\subsection*{Model transfer}
For the transfer and rescue experiments (\Cref{subsec:rescuing protocol}), we inject reasoning traces from one model into another model's context. This cross-model transfer requires handling potential tokenization mismatches between model families.

\paragraph{Text-level transfer.} Rather than transferring raw token IDs (which would be meaningless across different tokenizers), we operate at the text level. Reasoning traces are stored as decoded text strings, and the target model re-tokenizes this text using its own tokenizer. This approach ensures that each model processes the reasoning content using its native vocabulary and tokenization scheme.

\paragraph{Tag normalization.} When transferring traces between model families, we convert model-specific reasoning markers to the target model's expected format. Specifically, when injecting Qwen3 traces into gpt-oss models, we replace Qwen's \texttt{<think>} and \texttt{</think>} tags with the gpt-oss analysis channel tokens. Transfers within the same model family (e.g., Qwen3-4B to Qwen3-14B) require no tag conversion as both models share the same tokenizer and format conventions.

\paragraph{Decile computation and token length recalculation.} Deciles are defined with respect to the source model's tokenization: a decile-$d$ prefix consists of the first $\lceil d/100 \cdot L_{\text{source}} \rceil$ tokens from the source trace, decoded to text. This text slice is transferred as-is to the target model. After tag normalization, the target model re-tokenizes the transferred text using its own tokenizer, which may yield a slightly different token count due to vocabulary differences. The recalculated target-token length is used to verify context window constraints, not to redefine decile boundaries.

\paragraph{Free continuation.} The ``free continuation'' condition allows the target model to extend reasoning beyond the injected decile prefix before answering. We set \texttt{max\_tokens} to the remaining context budget (model context length minus prompt length minus a 128-token safety margin). Generation stops when either the token limit is reached or the model produces an end-of-sequence token. The model's extended reasoning is then measured using the same decile probing parameters described above.

\subsection{McNemar's test}
We use McNemar's test~\citep{McNemar_1947} to assess the statistical significance of accuracy differences between paired conditions (e.g., original traces vs.\ controls, or decile 0 vs.\ decile 100). McNemar's test is appropriate for comparing two related samples on a dichotomous outcome, like in our case, whether each question is answered correctly or incorrectly under two different conditions.

Given paired binary outcomes $(a_i, b_i)$ for each question $i$, where $a_i, b_i \in \{0, 1\}$ indicate correctness under conditions A and B respectively, McNemar's test focuses on the discordant pairs: questions where the two conditions disagree. Let $n_{10}$ denote the number of questions correct under A but incorrect under B, and $n_{01}$ the number incorrect under A but correct under B. The test statistic is:
\[
\chi^2 = \frac{(n_{01} - n_{10})^2}{n_{01} + n_{10}}
\]
which follows a chi-squared distribution with 1 degree of freedom under the null hypothesis that the two conditions have equal accuracy. For small samples ($n_{01} + n_{10} < 25$), we use the exact binomial test instead. We report significance levels as $^{*}p<0.05$, $^{**}p<0.01$, $^{***}p<0.001$.

\subsection{Hardware specifications}

Our experiments were performed on a HPC server with nodes containing 1 NVIDIA H200 GPU and system memory up to 256GB. We used vLLM 0.12.0 as inference engine. A small number of jobs ran on a CPU-only partition. The HPC experiments took approximately $1500$ GPU hours, and $80$ CPU hours. Omni-MATH-2 generations were judged via the OpenAI Batch API using GPT-5 mini.  

\subsection{Impact statement}
Our findings have practical implications for the deployment and monitoring of reasoning models. The observation that accuracy gains derive from instance-aligned semantic content (rather than mere length or structural form) suggests that naive truncation or trace-compression strategies risk discarding signal that materially affects correctness, informing work on token-efficient reasoning~\citep{aytes2025sketch,fan2025cothink}. At the same time, the presence of ``lost'' trajectories (7--11\% on GPQA Diamond) and the lack of improved confidence discrimination with reasoning depth indicate that longer reasoning does not guarantee better calibration and can produce highly confident wrong answers~\citep{phan2025humanity,lyu2025calibrating,marjanovic2025deepseek,wei2024measuring}. The trajectory-probing protocol also enables detection of model-specific failure modes that would remain hidden in aggregate evaluations: for instance, the Qwen3-8B \texttt{\textbackslash boxed\{\}} formatting anomaly (where the model frequently generates LaTeX boxing, biasing predictions toward ``A'') was only revealed by tracking per-decile accuracy, manifesting as a unique accuracy drop from decile 90 to 100. For multi-model pipelines, the rescue and anchoring results suggest that injecting partial traces across models can transfer useful but biasing state: stronger models benefit from early recovery opportunities but become increasingly anchored as traces grow longer. These dynamics inform design choices around when to terminate reasoning~\citep{liu-wang-2025-answer,sun2025stop,fu2025deep}, how to monitor intermediate states for reliability signals~\citep{baker2025monitoring,guan2025monitoring,korbak2025chain}, and whether to allow continuation versus immediate commitment when reusing traces across models~\citep{su2025between,stechly2025beyond,liu2025thought}.

\section{Supplementary Figures}
\begin{figure}[h]
    \centering
    \includegraphics[width=1\textwidth]{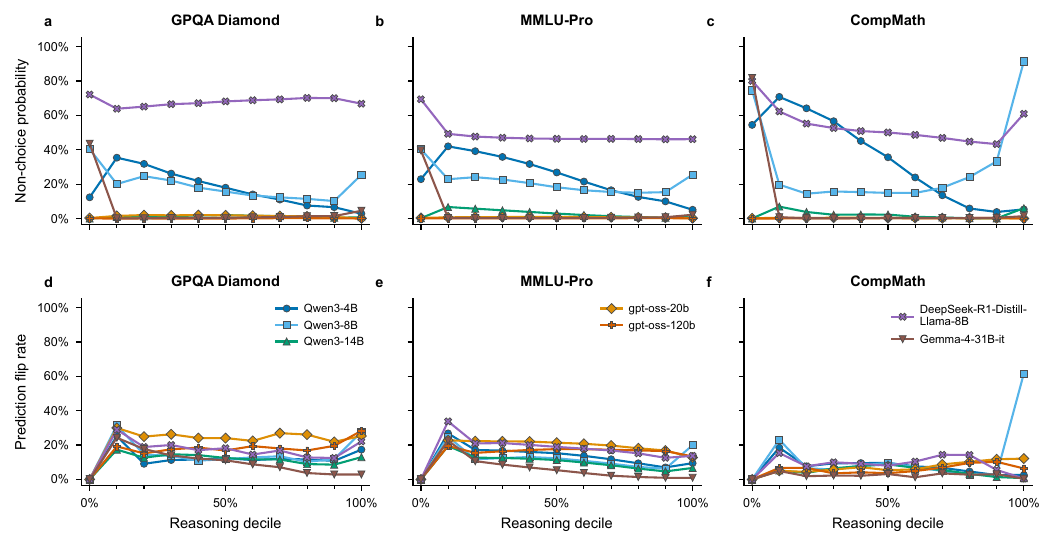}
\caption{Non-choice probability and flip rate on multiple-choice tasks. Companion to \Cref{fig:main_2}, covering seven models and three runs. Columns show GPQA Diamond, MMLU-Pro, and CompMath. \textbf{a--c}, Probability mass outside the answer-letter set. \textbf{d--f}, Fraction of predictions whose argmax answer differs from the preceding decile. Qwen3-8B exhibits an endpoint increase in non-choice mass and switching, most prominently on CompMath; DeepSeek-R1-Distill-Llama-8B also assigns substantial mass to non-choice tokens. These choice-based metrics are not reported for Omni-MATH-2.}
    \label{fig:appendix_main2_extra}
\end{figure}

\clearpage

  \begin{figure}[t]                                                   
      \centering
      \includegraphics[width=1\textwidth]{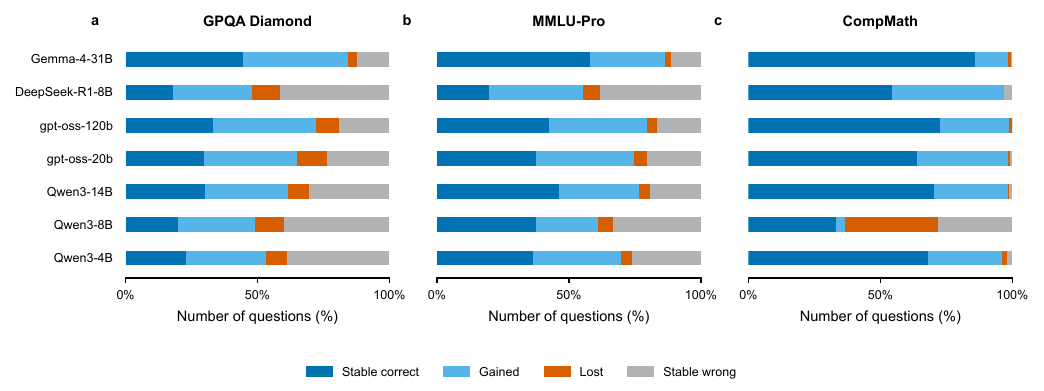}                     
\caption{Endpoint trajectory outcomes on the multiple-choice benchmarks. Stacked bars compare correctness at decile 0 and decile 100 for seven models on GPQA Diamond (\textbf{a}), MMLU-Pro (\textbf{b}), and CompMath (\textbf{c}), pooling runs 1--3. Stable correct (dark blue) denotes correct answers at both endpoints; gained (light blue), incorrect to correct; lost (orange), correct to incorrect; and stable wrong (gray), incorrect at both endpoints. Gained outcomes exceed lost outcomes for every model--benchmark combination except Qwen3-8B on CompMath, which has 35.2\% lost and 3.1\% gained outcomes.}
      \label{fig:appendix_trajectories}     
  \end{figure} 

  \begin{figure}[t]
    \centering
    \includegraphics[width=1\textwidth]{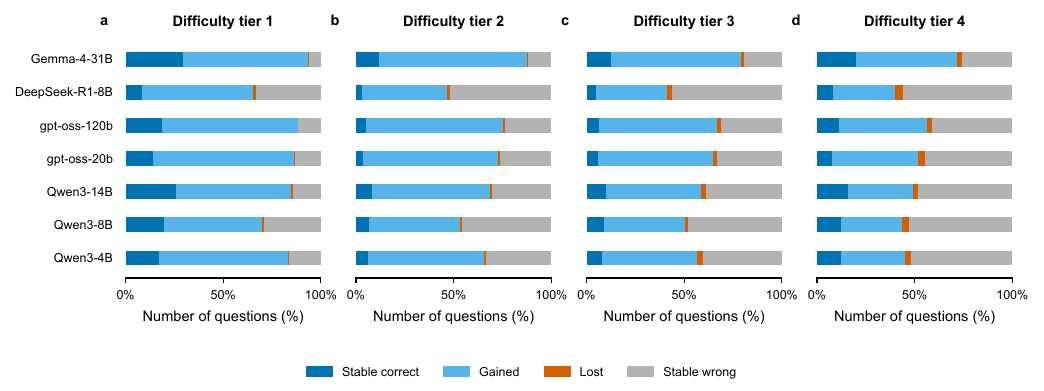}
\caption{Endpoint trajectory outcomes on Omni-MATH-2 by difficulty. Stacked bars compare judge-scored correctness at decile 0 and decile 100 for all seven models, pooling runs 1--3 within each difficulty tier. Panels \textbf{a--d} show increasingly difficult tiers formed from quartile cut points of the dataset's difficulty ratings. Colors denote stable correct, gained, lost, and stable wrong outcomes as in \Cref{fig:appendix_trajectories}. The figure separates net accuracy gains from individual correct-to-incorrect reversals within each tier.}
    \label{fig:appendix_omnimath_trajectories}
\end{figure}

  \begin{figure}[t]                                                   \centering                                                        \includegraphics[width=1\textwidth]{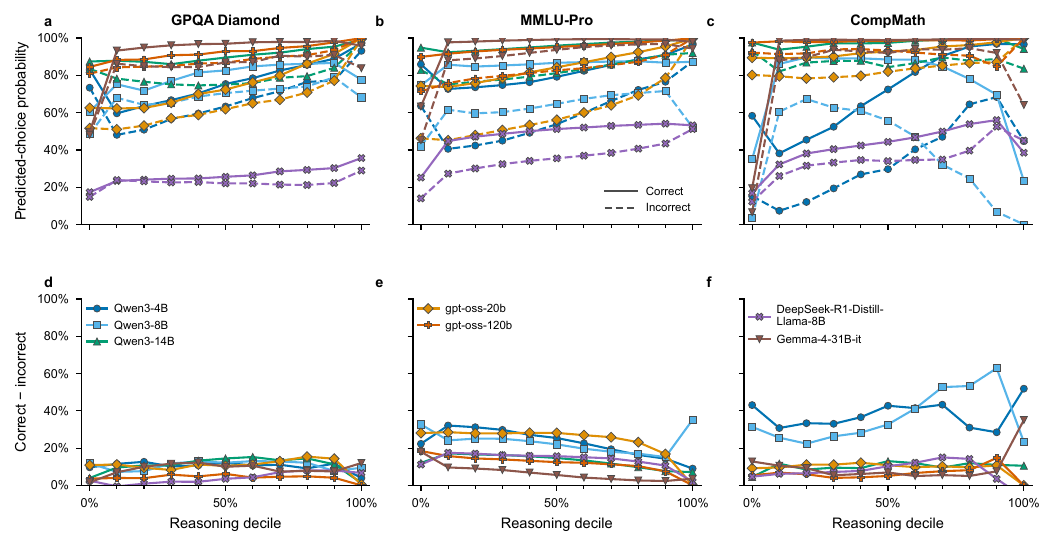} 
\caption{Prediction confidence for correct and incorrect decile predictions. Columns show GPQA Diamond, MMLU-Pro, and CompMath for all seven models. \textbf{a--c}, Mean unconditional probability assigned to the current argmax answer, separately for correct predictions (solid) and incorrect predictions (dashed). \textbf{d--f}, The correct-minus-incorrect prediction-confidence gap, in percentage points. Conditional means are computed within each run and then averaged over three runs. Here, prediction confidence is derived from answer-token probabilities; decision commitment in \Cref{fig:main_2} instead tracks the answer selected at the final decile. The gap measures discrimination between correct and incorrect predictions, not calibration. Overall, prediction confidence often rises for both correct and incorrect answers, while the gap does not consistently widen and narrows for most models on MMLU-Pro, indicating that additional reasoning does not consistently improve discrimination and can yield incorrect answers assigned high probability.}
    \label{fig:appendix_confidence}                 
  \end{figure}

    \begin{figure}[t]             
      \centering                                                    
      \includegraphics[width=1\textwidth]{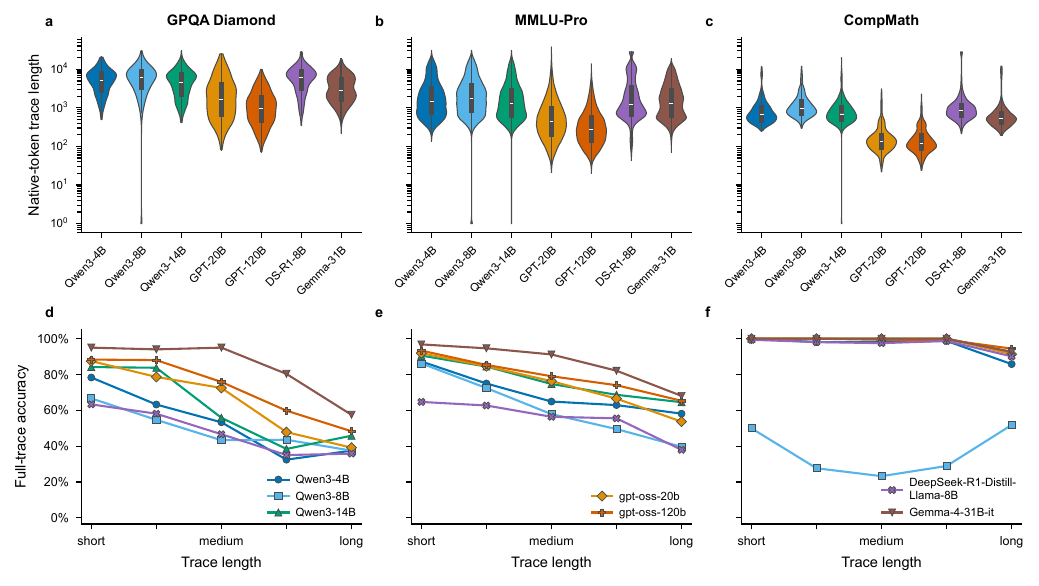}             
      \caption{Reasoning trace length distributions and their relationship to accuracy. Analysis of full reasoning traces (decile 100) for the seven models on GPQA Diamond, MMLU-Pro, and CompMath, averaged over 3 runs. \textbf{a,b}, Violin plots showing trace length distributions (in tokens) by model on GPQA Diamond and MMLU-Pro. gpt-oss models produce substantially shorter traces than Qwen3 models, with gpt-oss-120b generating the most concise reasoning. \textbf{c,d}, Accuracy stratified by trace-length quintile (shortest to longest) reveals a consistent negative relationship on GQPA Diamond and MMLU-Pro: shorter traces are associated with higher accuracy across all models and benchmarks. This pattern suggests models produce extended reasoning when struggling with difficult questions, rather than extended reasoning leading to correct answers. On CompMath, six models remain near ceiling through the first four bins and have lower accuracy in the longest bin; Qwen3-8B instead shows a non-monotonic pattern. }                
      \label{fig:appendix_accuracy_length} 
  \end{figure}

  \begin{figure}[t]
    \centering
    \includegraphics[width=1\textwidth]{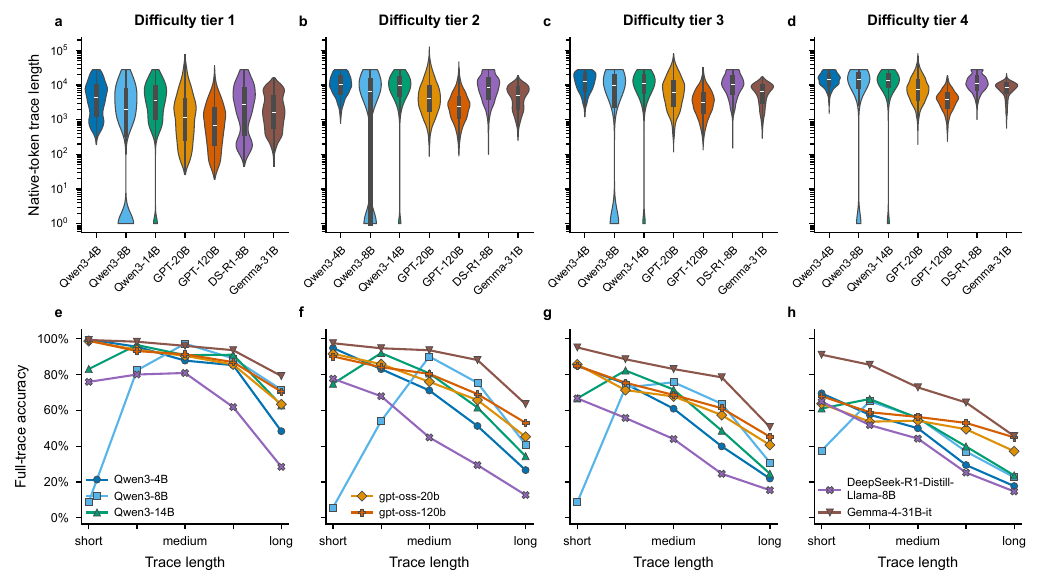}
\caption{Reasoning trace lengths and endpoint accuracy on Omni-MATH-2, stratified by difficulty. Columns show four increasingly difficult tiers for all seven models and three runs. \textbf{a--d}, Distributions of reasoning trace length in native tokens, on logarithmic axes. \textbf{e--h}, Judge-scored decile-100 accuracy across length quintiles formed within each model, run, and difficulty tier, averaged over runs. The overall negative association between trace length and accuracy persists within difficulty tiers for the six models without the Qwen3-8B endpoint anomaly; Qwen3-8B shows a non-monotonic relationship.}
    \label{fig:appendix_omnimath_accuracy_length}
\end{figure}

\clearpage

\begin{figure}[t]
    \centering
    \includegraphics[width=1\textwidth]{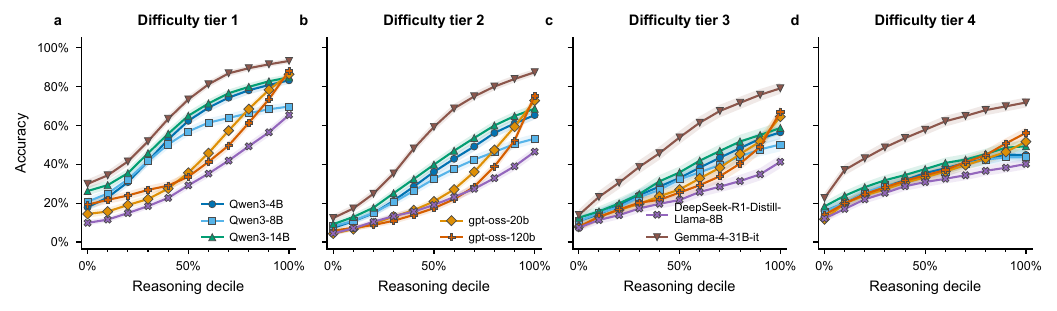}
    \caption{Accuracy on Omni-MATH-2 at each reasoning decile, stratified by the dataset's predefined difficulty rating (1--9.5). The increase persists within every quartile and every model, ruling out question difficulty as a driver of the trajectory dynamics. Averaged over three runs.
    }
    \label{fig:appendix_omnimath_difficulty}
\end{figure}

  \begin{figure}[t]
    \centering
    \includegraphics[width=1\textwidth]{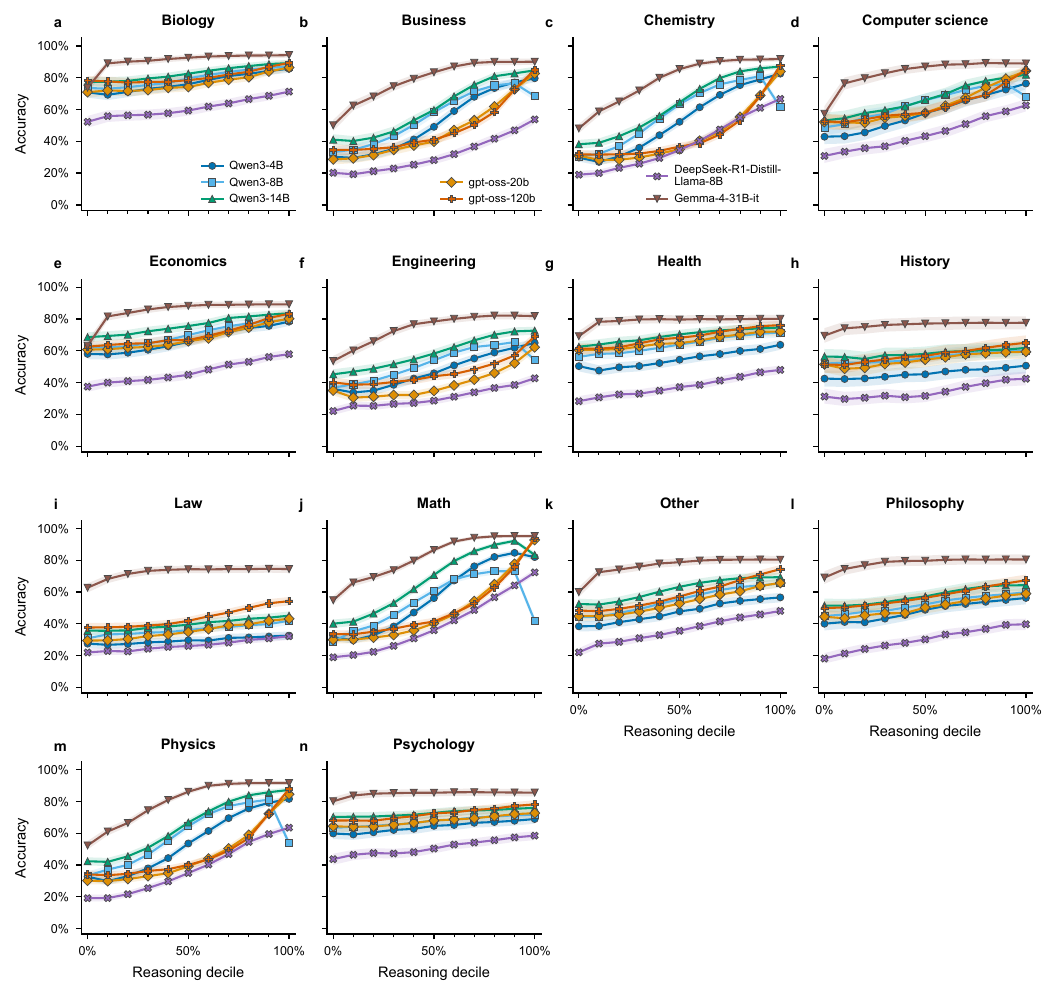}
    \caption{Accuracy on MMLU-Pro at each reasoning decile, stratified across the 14 MMLU-Pro subject categories. The accuracy increase holds universally with the exception of Qwen3-8B (see \Cref{tab:boxed-collapse-full}); reasoning gains are steeper on subjects like Math and Physics than on subjects like Law and Psychology. Averaged over 3 runs.}
    \label{fig:appendix_stratified_mmlu_categories}
\end{figure}

\begin{figure}[t]
    \centering
    \includegraphics[width=1\textwidth]{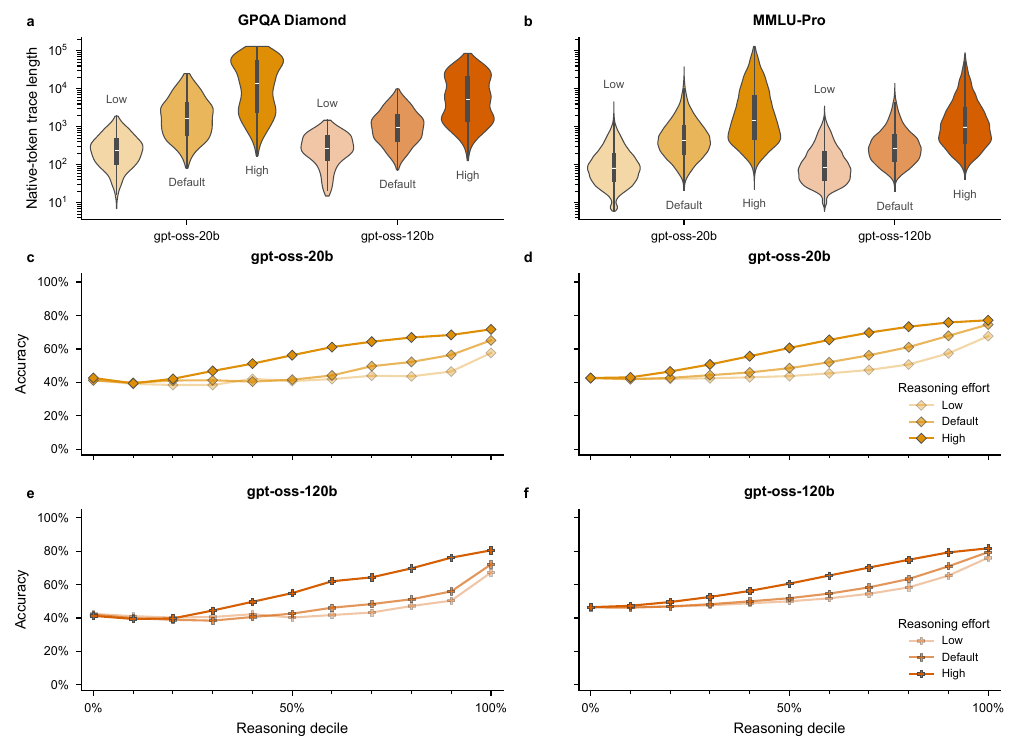}
    \caption{Per-decile accuracy at three reasoning-effort settings (gpt-oss family). Accuracy at each reasoning decile for gpt-oss-20b and gpt-oss-120b on GPQA Diamond and MMLU-Pro, at low (light), medium (mid), and high (dark) reasoning effort. Low-effort traces plateau earlier; high-effort traces continue to gain in the second half of the trace. Averaged over 3 runs.}
    \label{fig:appendix_effort}
\end{figure}

\begin{figure}[t]
    \centering
    \includegraphics[width=1\textwidth]{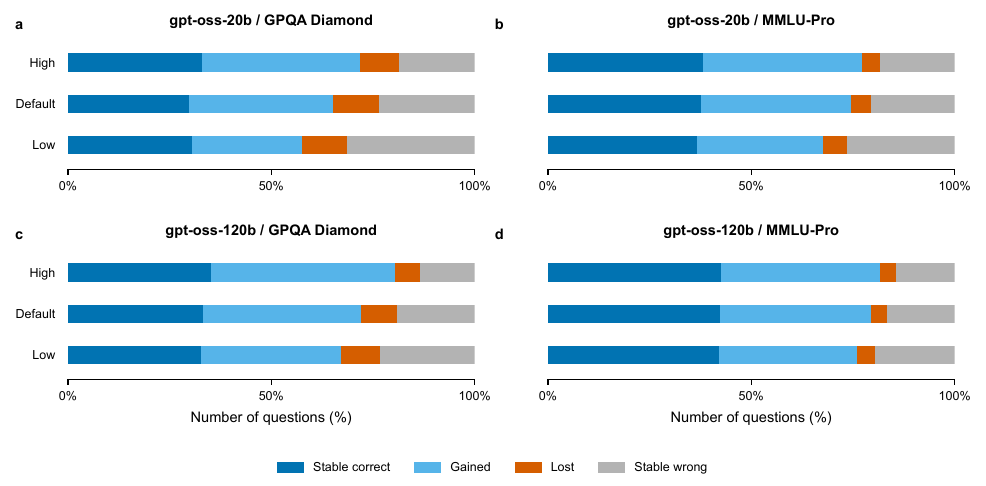}
    \caption{Trajectory outcomes by reasoning-effort setting. Stacked bar charts of decile-0 vs.\ decile-100 outcomes (stable correct / gained / lost / stable wrong) per reasoning effort level for gpt-oss-20b and gpt-oss-120b on GPQA Diamond and MMLU-Pro. The gained fraction is largest at high effort; the lost fraction is approximately stable across effort levels. Averaged over 3 runs.}
    \label{fig:appendix_effort_trajectories}
\end{figure}

\clearpage 

\begin{figure}[t]
    \centering
    \includegraphics[width=1\textwidth]{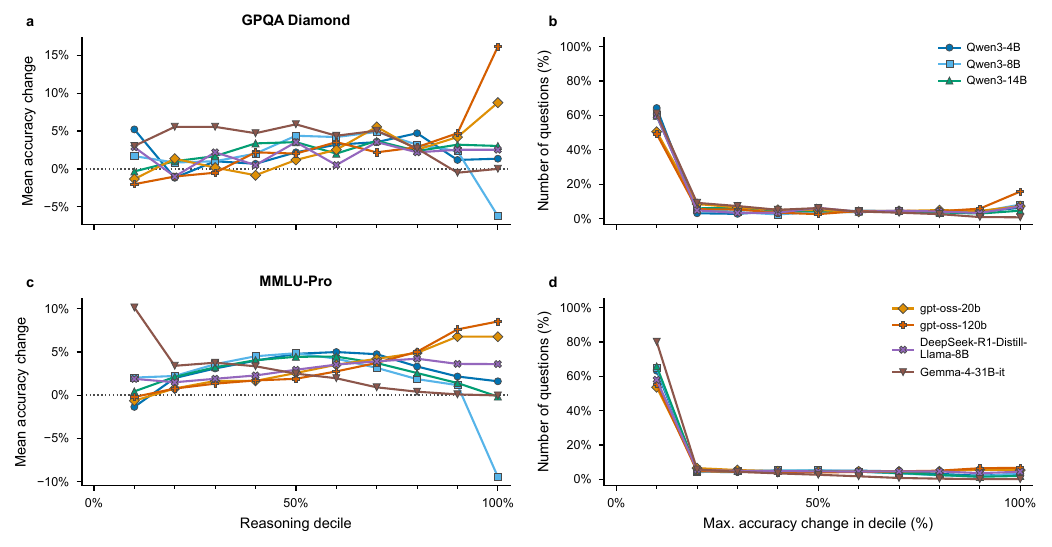}
    \caption{Distribution of accuracy gains across deciles. Post-hoc analysis of where accuracy gains concentrate within the reasoning trajectory. \textbf{a,c}, Mean accuracy change per consecutive decile step on GPQA Diamond and MMLU-Pro. Gains are spread approximately uniformly across all decile intervals, with no single step dominating. \textbf{b,d}, Distribution of the per-question peak-change decile. Beyond the initial $0\%\rightarrow 10\%$ step, peak changes are spread approximately uniformly across deciles. This validates the fixed 10\% decile slicing as a sensible default; semantic-boundary truncation~\citep{muennighoff2025s1simpletesttimescaling} is complementary, providing finer model-specific resolution.}
    \label{fig:appendix_change_points}
\end{figure}

\begin{figure}[t]
    \centering
    \includegraphics[width=1\textwidth]{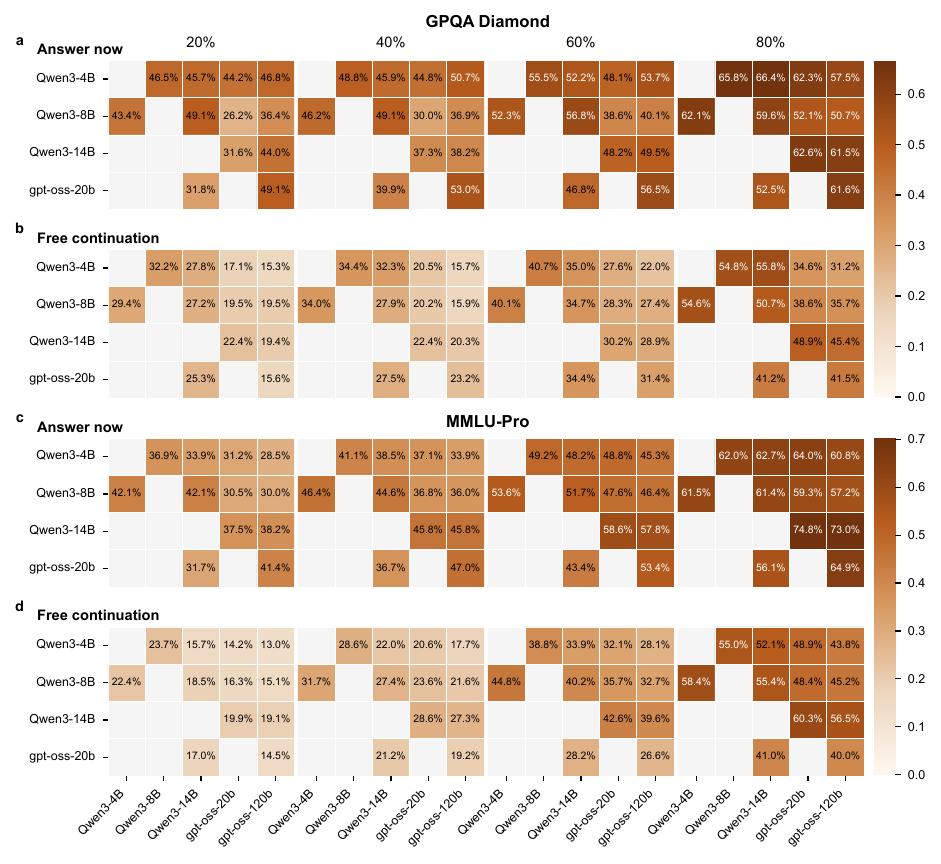}
 \caption{Anchoring rates in cross-model probing. Each cell reports the percentage of selected transfers for which the target repeats the source model's incorrect answer. Layout and cohorts match \Cref{fig:main_4}, with ten forward and two reverse model pairs: \textbf{a,b}, GPQA Diamond, pooling runs 1--3; \textbf{c,d}, MMLU-Pro, run 1; answer now in \textbf{a,c} and free continuation in \textbf{b,d}. Within each target-model block, columns show 20\%, 40\%, 60\%, and 80\% prefixes. Blank cells denote unmeasured pairs; two context-limited MMLU-Pro reverse measurements are omitted. Free continuation reduces anchoring, but substantial anchoring remains at later prefixes.}
    \label{fig:appendix_anchoring}
\end{figure}

\begin{figure}
    \centering
    \includegraphics[width=\linewidth]{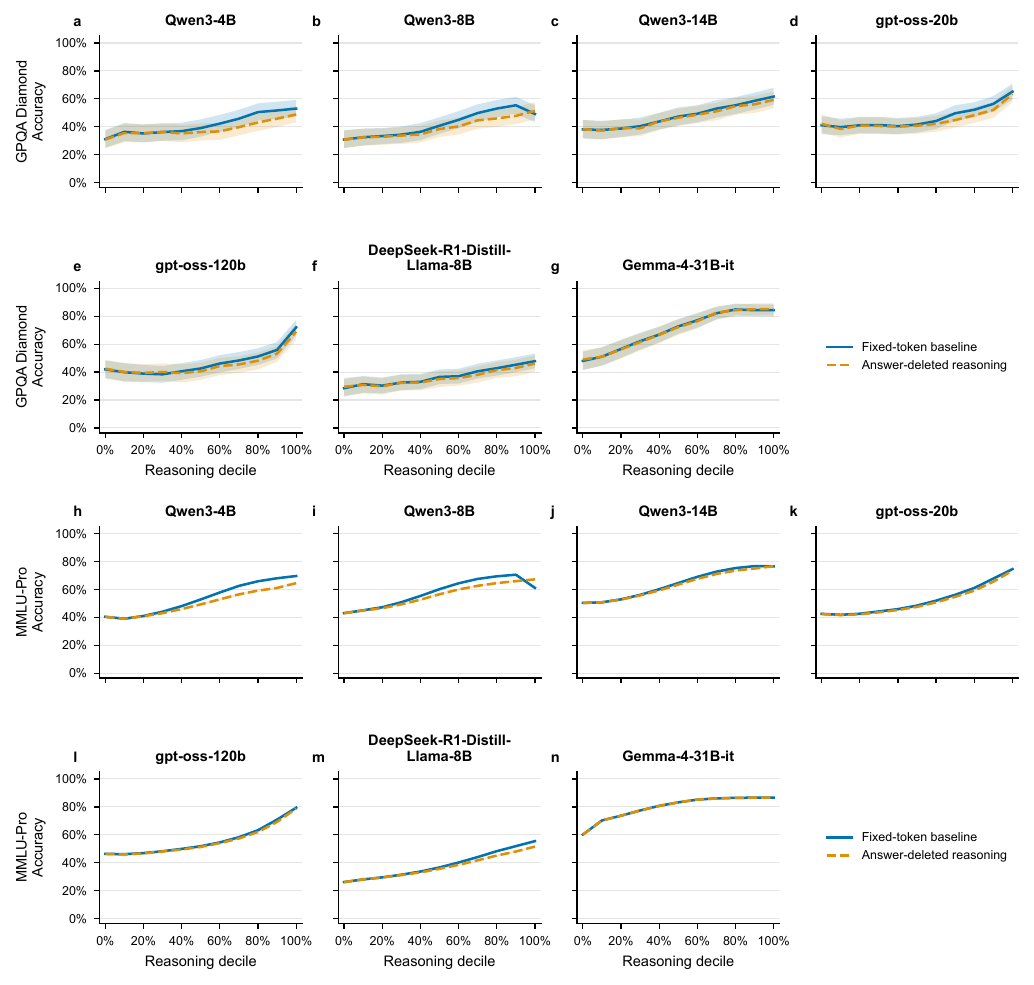}
 \caption{Accuracy with answer-deleted reasoning prefixes. Fixed-token baseline (solid blue) versus the answer-deleted reasoning condition (dashed orange) for seven models on GPQA Diamond (\textbf{a--g}) and MMLU-Pro (\textbf{h--n}). Curves average three runs; shading shows pointwise 95\% question-bootstrap confidence intervals (2{,}000 resamples, retaining paired conditions and runs together). Broad accuracy gains remain, but the intervention changes some trajectories and endpoints, including Qwen3-8B's endpoint behavior.}
    \label{fig:placeholder}
\end{figure}

\begin{figure}
    \centering
    \includegraphics[width=\linewidth]{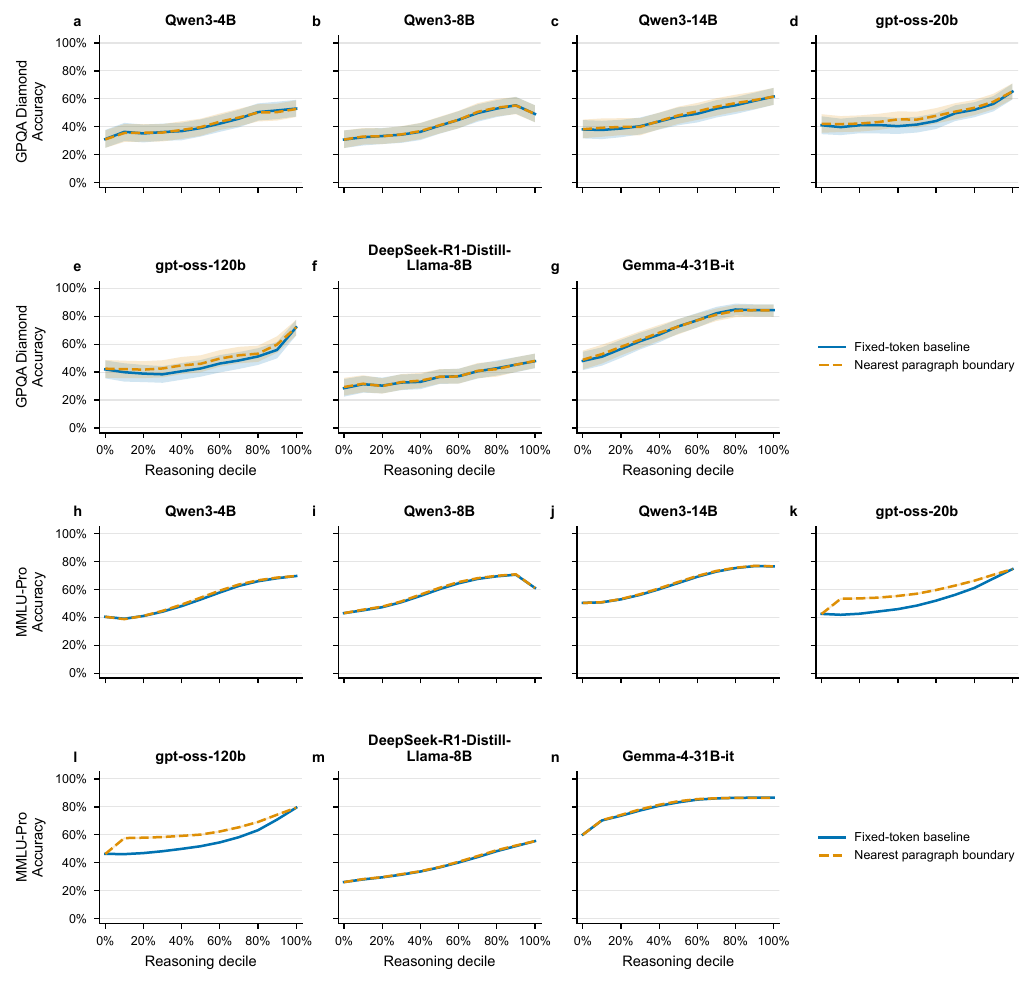}
\caption{Sensitivity to truncation at paragraph boundaries. Fixed-token deciles (solid blue) versus the nearest-paragraph-boundary condition (dashed orange) for seven models on GPQA Diamond (\textbf{a--g}) and MMLU-Pro (\textbf{h--n}). Curves average three runs; shading shows pointwise 95\% question-bootstrap confidence intervals (2{,}000 resamples, retaining paired conditions and runs together). The horizontal axis denotes the requested reasoning decile. Overall patterns are similar, with visible intermediate-decile differences, especially for the gpt-oss models on MMLU-Pro.}
    \label{fig:placeholder}
\end{figure}
\clearpage

\section{Supplementary tables}
\begin{table}[!ht]
\centering
\caption{Run-to-run variability at full reasoning. Mean $\pm$ sample standard deviation over three runs at 100\% reasoning. Values are percentages. For multiple-choice datasets, commitment is the probability assigned to the final predicted answer; non-choice probability is one minus the total answer-choice mass. For Omni-MATH-2, confidence is the mean generated-answer token probability, not a probability over answer choices; non-choice mass and multiple-choice flip rate are not applicable. Variation across runs is distinct from the question-bootstrap confidence intervals in the figures.}
\label{tab:appendix_variance}
\begingroup
\small
\setlength{\tabcolsep}{3pt}
\renewcommand{\arraystretch}{1.12}
\sbox{0}{%
\begin{tabular}{@{}lrrrr@{}}
\toprule
Model & Accuracy & \shortstack{Commitment /\\confidence} & \shortstack{Non-choice\\probability} & \shortstack{Flip rate\\90--100\%} \\
\midrule
\multicolumn{5}{@{}l}{\textbf{GPQA Diamond ($n=198$ questions per run)}} \\
\addlinespace[2pt]
Qwen3-4B & $53.0 \pm 2.7$ & $95.3 \pm 0.4$ & $2.8 \pm 0.3$ & $17.3 \pm 0.6$ \\
Qwen3-8B & $49.2 \pm 1.1$ & $72.8 \pm 0.7$ & $25.4 \pm 1.3$ & $27.6 \pm 1.8$ \\
Qwen3-14B & $61.6 \pm 1.3$ & $98.7 \pm 0.3$ & $0.7 \pm 0.6$ & $13.0 \pm 1.1$ \\
gpt-oss-20b & $65.2 \pm 2.3$ & $99.8 \pm 0.2$ & $0.1 \pm 0.1$ & $25.3 \pm 1.3$ \\
gpt-oss-120b & $72.1 \pm 2.5$ & $100.0 \pm 0.0$ & $0.0 \pm 0.0$ & $28.3 \pm 2.5$ \\
DeepSeek-R1-Distill-Llama-8B & $47.8 \pm 2.9$ & $32.1 \pm 0.5$ & $66.8 \pm 0.4$ & $22.1 \pm 1.6$ \\
Gemma-4-31B-it & $84.3 \pm 2.7$ & $94.0 \pm 0.7$ & $4.5 \pm 0.7$ & $2.9 \pm 1.1$ \\
\midrule
\multicolumn{5}{@{}l}{\textbf{MMLU-Pro ($n=12,032$ questions per run)}} \\
\addlinespace[2pt]
Qwen3-4B & $69.7 \pm 0.1$ & $93.8 \pm 0.0$ & $5.1 \pm 0.1$ & $9.5 \pm 0.2$ \\
Qwen3-8B & $61.1 \pm 0.2$ & $73.6 \pm 0.1$ & $25.5 \pm 0.1$ & $20.0 \pm 0.1$ \\
Qwen3-14B & $76.6 \pm 0.1$ & $97.5 \pm 0.0$ & $1.9 \pm 0.0$ & $6.7 \pm 0.2$ \\
gpt-oss-20b & $74.6 \pm 0.1$ & $99.8 \pm 0.1$ & $0.2 \pm 0.1$ & $12.7 \pm 0.4$ \\
gpt-oss-120b & $79.4 \pm 0.1$ & $99.9 \pm 0.0$ & $0.1 \pm 0.0$ & $13.4 \pm 0.3$ \\
DeepSeek-R1-Distill-Llama-8B & $55.5 \pm 0.2$ & $52.3 \pm 0.1$ & $46.2 \pm 0.1$ & $13.7 \pm 0.2$ \\
Gemma-4-31B-it & $86.6 \pm 0.0$ & $97.5 \pm 0.1$ & $2.2 \pm 0.1$ & $0.9 \pm 0.1$ \\
\midrule
\multicolumn{5}{@{}l}{\textbf{CompMath ($n=267$ questions per run)}} \\
\addlinespace[2pt]
Qwen3-4B & $96.1 \pm 0.6$ & $94.6 \pm 1.3$ & $5.3 \pm 1.3$ & $2.6 \pm 1.4$ \\
Qwen3-8B & $36.5 \pm 0.6$ & $8.5 \pm 0.3$ & $91.5 \pm 0.3$ & $61.7 \pm 1.1$ \\
Qwen3-14B & $98.5 \pm 0.4$ & $94.0 \pm 0.7$ & $6.0 \pm 0.7$ & $0.7 \pm 0.6$ \\
gpt-oss-20b & $98.3 \pm 0.2$ & $100.0 \pm 0.0$ & $0.0 \pm 0.0$ & $12.1 \pm 0.2$ \\
gpt-oss-120b & $98.6 \pm 0.4$ & $100.0 \pm 0.0$ & $0.0 \pm 0.0$ & $6.4 \pm 1.7$ \\
DeepSeek-R1-Distill-Llama-8B & $96.8 \pm 0.9$ & $38.6 \pm 3.9$ & $61.0 \pm 3.9$ & $0.5 \pm 0.2$ \\
Gemma-4-31B-it & $98.5 \pm 0.4$ & $98.6 \pm 0.1$ & $1.3 \pm 0.1$ & $0.4 \pm 0.4$ \\
\midrule
\multicolumn{5}{@{}l}{\textbf{Omni-MATH-2 ($n=4,181$ questions per run)}} \\
\addlinespace[2pt]
Qwen3-4B & $65.3 \pm 0.3$ & $80.6 \pm 1.2$ & -- & -- \\
Qwen3-8B & $56.0 \pm 0.3$ & $78.5 \pm 0.1$ & -- & -- \\
Qwen3-14B & $68.1 \pm 0.3$ & $86.4 \pm 0.2$ & -- & -- \\
gpt-oss-20b & $71.3 \pm 1.3$ & $69.6 \pm 0.2$ & -- & -- \\
gpt-oss-120b & $74.0 \pm 0.5$ & $80.7 \pm 0.2$ & -- & -- \\
DeepSeek-R1-Distill-Llama-8B & $50.5 \pm 0.3$ & $76.6 \pm 3.5$ & -- & -- \\
Gemma-4-31B-it & $84.6 \pm 0.2$ & $96.1 \pm 0.1$ & -- & -- \\
\bottomrule
\end{tabular}}
\ifdim\wd0>\linewidth\resizebox{\linewidth}{!}{\usebox{0}}\else\usebox{0}\fi

\endgroup
\end{table}

\begin{table}[p]
\centering
\caption{Accuracy gains and control comparisons. Accuracy at 0\% and 100\% reasoning; gains and original-minus-control differences are in percentage points. All cells use runs 1--3. $n$ counts paired question--run observations (three per question). Controls are evaluated at 100\% reasoning. $^{*}p<0.05$, $^{**}p<0.01$, $^{***}p<0.001$: two-sided McNemar tests, exact for fewer than 25 discordant pairs and continuity-corrected otherwise. These pooled tests retain the original analysis convention and do not adjust for dependence between repeated runs of a question. No corresponding control experiment was run for open-ended Omni-MATH-2.}
\label{tab:summary-statistics}
\begingroup
\small
\setlength{\tabcolsep}{3pt}
\renewcommand{\arraystretch}{1.12}
\sbox{0}{%
\begin{tabular}{@{}lrrrrrrr@{}}
\toprule
 & & \multicolumn{2}{c}{Accuracy (\%)} & & \multicolumn{3}{c}{Original $-$ control (pp)} \\
\cmidrule(lr){3-4}\cmidrule(lr){6-8}
Model & $n$ & 0\% & 100\% & Gain (pp) & Random & Swap & Shuffle \\
\midrule
\multicolumn{8}{@{}l}{\textbf{GPQA Diamond}} \\
\addlinespace[2pt]
Qwen3-4B & 594 & 31.1 & 53.0 & $+21.9^{***}$ & $+19.0^{***}$ & $+19.0^{***}$ & $+12.8^{***}$ \\
Qwen3-8B & 594 & 30.8 & 49.2 & $+18.4^{***}$ & $+14.8^{***}$ & $+15.7^{***}$ & $+14.0^{***}$ \\
Qwen3-14B & 594 & 38.0 & 61.6 & $+23.6^{***}$ & $+22.6^{***}$ & $+22.7^{***}$ & $+15.7^{***}$ \\
gpt-oss-20b & 594 & 41.1 & 65.2 & $+24.1^{***}$ & $+25.3^{***}$ & $+28.8^{***}$ & $+21.5^{***}$ \\
gpt-oss-120b & 594 & 41.9 & 72.1 & $+30.1^{***}$ & $+33.5^{***}$ & $+33.5^{***}$ & $+29.1^{***}$ \\
DeepSeek-R1-Distill-Llama-8B & 594 & 28.5 & 47.8 & $+19.4^{***}$ & $+16.3^{***}$ & $+18.2^{***}$ & $+16.0^{***}$ \\
Gemma-4-31B-it & 594 & 48.0 & 84.3 & $+36.4^{***}$ & $+34.2^{***}$ & $+44.3^{***}$ & $+31.8^{***}$ \\
\midrule
\multicolumn{8}{@{}l}{\textbf{MMLU-Pro}} \\
\addlinespace[2pt]
Qwen3-4B & 36,096 & 40.4 & 69.7 & $+29.2^{***}$ & $+29.6^{***}$ & $+35.5^{***}$ & $+22.5^{***}$ \\
Qwen3-8B & 36,096 & 43.1 & 61.1 & $+18.1^{***}$ & $+17.3^{***}$ & $+25.2^{***}$ & $+13.3^{***}$ \\
Qwen3-14B & 36,096 & 50.4 & 76.6 & $+26.2^{***}$ & $+27.0^{***}$ & $+40.3^{***}$ & $+20.0^{***}$ \\
gpt-oss-20b & 36,096 & 42.5 & 74.6 & $+32.1^{***}$ & $+33.1^{***}$ & $+50.5^{***}$ & $+21.6^{***}$ \\
gpt-oss-120b & 36,096 & 46.4 & 79.4 & $+33.1^{***}$ & $+34.7^{***}$ & $+46.3^{***}$ & $+22.3^{***}$ \\
DeepSeek-R1-Distill-Llama-8B & 36,096 & 26.2 & 55.5 & $+29.3^{***}$ & $+28.4^{***}$ & $+36.5^{***}$ & $+23.7^{***}$ \\
Gemma-4-31B-it & 36,096 & 60.2 & 86.6 & $+26.4^{***}$ & $+19.9^{***}$ & $+40.9^{***}$ & $+16.9^{***}$ \\
\midrule
\multicolumn{8}{@{}l}{\textbf{CompMath}} \\
\addlinespace[2pt]
Qwen3-4B & 801 & 69.9 & 96.1 & $+26.2^{***}$ & $+25.0^{***}$ & $+35.5^{***}$ & $+16.9^{***}$ \\
Qwen3-8B & 801 & 68.5 & 36.5 & $-32.1^{***}$ & $-32.8^{***}$ & $-21.8^{***}$ & $-41.6^{***}$ \\
Qwen3-14B & 801 & 70.8 & 98.5 & $+27.7^{***}$ & $+25.1^{***}$ & $+47.7^{***}$ & $+20.8^{***}$ \\
gpt-oss-20b & 801 & 64.7 & 98.3 & $+33.6^{***}$ & $+36.1^{***}$ & $+46.2^{***}$ & $+25.6^{***}$ \\
gpt-oss-120b & 801 & 73.9 & 98.6 & $+24.7^{***}$ & $+29.5^{***}$ & $+35.6^{***}$ & $+21.8^{***}$ \\
DeepSeek-R1-Distill-Llama-8B & 801 & 54.7 & 96.8 & $+42.1^{***}$ & $+42.3^{***}$ & $+54.9^{***}$ & $+23.8^{***}$ \\
Gemma-4-31B-it & 801 & 86.9 & 98.5 & $+11.6^{***}$ & $+10.9^{***}$ & $+26.3^{***}$ & $+10.0^{***}$ \\
\bottomrule
\end{tabular}}
\ifdim\wd0>\linewidth\resizebox{\linewidth}{!}{\usebox{0}}\else\usebox{0}\fi

\endgroup
\end{table}

\begin{table}[p]
\centering
\caption{Boxed-format occurrence and endpoint predictions. Full-response formatting and 100\%-reasoning predictions, pooled over runs 1--3. Rates and accuracies are percentages. Boxed denotes the literal, case-sensitive substring \texttt{\textbackslash boxed} anywhere in the full saved response, including reasoning and final reply; valid braces or a valid boxed answer are not required. $P(A)$ is the fraction of predictions selecting choice A within each subgroup. $n_{\rm boxed}$ counts question--run observations, not unique questions. Dashes denote an empty subgroup. These comparisons are descriptive associations.}
\label{tab:boxed-collapse-full}
\begingroup
\small
\setlength{\tabcolsep}{3pt}
\renewcommand{\arraystretch}{1.12}
\sbox{0}{%
\begin{tabular}{@{}lrrrrrr@{}}
\toprule
Model & $n_{\rm boxed}$ & Boxed (\%) & \shortstack{$P(A)$\\boxed} & \shortstack{$P(A)$\\non-boxed} & \shortstack{Accuracy\\boxed} & \shortstack{Accuracy\\non-boxed} \\
\midrule
\multicolumn{7}{@{}l}{\textbf{GPQA Diamond}} \\
\addlinespace[2pt]
Qwen3-4B & 5 & 0.8 & 100.0 & 21.7 & 0.0 & 53.5 \\
Qwen3-8B & 168 & 28.3 & 79.2 & 18.8 & 46.4 & 50.2 \\
Qwen3-14B & 4 & 0.7 & 75.0 & 23.2 & 50.0 & 61.7 \\
gpt-oss-20b & 0 & 0.0 & -- & 26.6 & -- & 65.2 \\
gpt-oss-120b & 0 & 0.0 & -- & 29.6 & -- & 72.1 \\
DeepSeek-R1-Distill-Llama-8B & 133 & 22.4 & 28.6 & 24.3 & 58.6 & 44.7 \\
Gemma-4-31B-it & 0 & 0.0 & -- & 28.6 & -- & 84.3 \\
\midrule
\multicolumn{7}{@{}l}{\textbf{MMLU-Pro}} \\
\addlinespace[2pt]
Qwen3-4B & 1,241 & 3.4 & 57.6 & 11.4 & 41.7 & 70.6 \\
Qwen3-8B & 9,914 & 27.5 & 68.3 & 11.4 & 33.6 & 71.6 \\
Qwen3-14B & 682 & 1.9 & 79.8 & 11.4 & 24.0 & 77.6 \\
gpt-oss-20b & 0 & 0.0 & -- & 11.5 & -- & 74.6 \\
gpt-oss-120b & 0 & 0.0 & -- & 11.7 & -- & 79.4 \\
DeepSeek-R1-Distill-Llama-8B & 6,627 & 18.4 & 11.9 & 12.0 & 61.6 & 54.1 \\
Gemma-4-31B-it & 0 & 0.0 & -- & 11.7 & -- & 86.6 \\
\midrule
\multicolumn{7}{@{}l}{\textbf{CompMath}} \\
\addlinespace[2pt]
Qwen3-4B & 33 & 4.1 & 72.7 & 29.2 & 60.6 & 97.7 \\
Qwen3-8B & 733 & 91.5 & 99.7 & 17.6 & 30.6 & 100.0 \\
Qwen3-14B & 49 & 6.1 & 40.8 & 28.9 & 95.9 & 98.7 \\
gpt-oss-20b & 0 & 0.0 & -- & 29.8 & -- & 98.3 \\
gpt-oss-120b & 0 & 0.0 & -- & 29.7 & -- & 98.6 \\
DeepSeek-R1-Distill-Llama-8B & 419 & 52.3 & 31.5 & 28.8 & 98.1 & 95.3 \\
Gemma-4-31B-it & 0 & 0.0 & -- & 28.8 & -- & 98.5 \\
\bottomrule
\end{tabular}}
\ifdim\wd0>\linewidth\resizebox{\linewidth}{!}{\usebox{0}}\else\usebox{0}\fi

\endgroup
\end{table}

\begin{table}[p]
\centering
\caption{Cross-model rescue and anchoring. Corrected forward-transfer cohorts, pooled over the measured 20\%, 40\%, 60\%, and 80\% prefixes. Rates are percentages; $\Delta$ is free continuation minus answer now, in percentage points. Rescue means answering correctly; anchoring means repeating the source model's wrong answer. $n$ counts measured question--run--decile transfers. Selection uses the original saved-score order to resolve exact ties. $^{*}p<0.05$, $^{**}p<0.01$, $^{***}p<0.001$: one-sided McNemar tests for increased rescue or decreased anchoring, using the original pooled convention without adjustment for repeated questions. \textit{Coverage:} the two reverse directions (Qwen3-8B$\to$Qwen3-4B and gpt-oss-20b$\to$Qwen3-14B) are excluded pending reconciliation with current source cohorts.}
\label{tab:rescue-summary}
\begingroup
\small
\setlength{\tabcolsep}{3pt}
\renewcommand{\arraystretch}{1.12}
\sbox{0}{%
\begin{tabular}{@{}lrrrrrrr@{}}
\toprule
 & & \multicolumn{3}{c}{Rescue} & \multicolumn{3}{c}{Anchoring} \\
\cmidrule(lr){3-5}\cmidrule(lr){6-8}
Source $\to$ target & $n$ & Now & Free & $\Delta$ & Now & Free & $\Delta$ \\
\midrule
\multicolumn{8}{@{}l}{\textbf{GPQA Diamond (runs 1--3)}} \\
\addlinespace[2pt]
Qwen3-4B$\to$Qwen3-8B & 1,389 & 20.4 & 36.1 & $+15.7^{***}$ & 53.3 & 39.6 & $-13.8^{***}$ \\
Qwen3-4B$\to$Qwen3-14B & 1,389 & 25.1 & 43.4 & $+18.4^{***}$ & 51.7 & 36.6 & $-15.0^{***}$ \\
Qwen3-4B$\to$gpt-oss-20b & 1,389 & 26.0 & 57.0 & $+31.0^{***}$ & 49.1 & 24.3 & $-24.8^{***}$ \\
Qwen3-4B$\to$gpt-oss-120b & 1,389 & 24.9 & 59.7 & $+34.8^{***}$ & 51.8 & 20.4 & $-31.4^{***}$ \\
Qwen3-8B$\to$Qwen3-14B & 1,387 & 23.7 & 44.7 & $+21.0^{***}$ & 53.1 & 33.9 & $-19.2^{***}$ \\
Qwen3-8B$\to$gpt-oss-20b & 1,387 & 28.9 & 55.9 & $+27.0^{***}$ & 35.4 & 25.6 & $-9.8^{***}$ \\
Qwen3-8B$\to$gpt-oss-120b & 1,387 & 29.2 & 58.0 & $+28.8^{***}$ & 40.3 & 23.6 & $-16.7^{***}$ \\
Qwen3-14B$\to$gpt-oss-20b & 1,259 & 23.9 & 53.3 & $+29.4^{***}$ & 43.5 & 29.8 & $-13.7^{***}$ \\
Qwen3-14B$\to$gpt-oss-120b & 1,259 & 25.2 & 56.6 & $+31.4^{***}$ & 47.4 & 27.3 & $-20.1^{***}$ \\
gpt-oss-20b$\to$gpt-oss-120b & 1,320 & 20.1 & 55.9 & $+35.8^{***}$ & 54.7 & 27.2 & $-27.5^{***}$ \\
\midrule
\multicolumn{8}{@{}l}{\textbf{MMLU-Pro (run 1)}} \\
\addlinespace[2pt]
Qwen3-4B$\to$Qwen3-8B & 22,515 & 21.7 & 38.0 & $+16.3^{***}$ & 45.4 & 34.2 & $-11.3^{***}$ \\
Qwen3-4B$\to$Qwen3-14B & 22,515 & 27.4 & 52.9 & $+25.5^{***}$ & 43.6 & 28.2 & $-15.5^{***}$ \\
Qwen3-4B$\to$gpt-oss-20b & 22,515 & 22.0 & 53.8 & $+31.8^{***}$ & 42.8 & 26.3 & $-16.4^{***}$ \\
Qwen3-4B$\to$gpt-oss-120b & 22,515 & 27.5 & 57.7 & $+30.2^{***}$ & 39.6 & 23.3 & $-16.3^{***}$ \\
Qwen3-8B$\to$Qwen3-14B & 19,605 & 25.7 & 49.2 & $+23.5^{***}$ & 48.5 & 32.5 & $-15.9^{***}$ \\
Qwen3-8B$\to$gpt-oss-20b & 19,605 & 23.3 & 52.4 & $+29.1^{***}$ & 41.3 & 28.5 & $-12.8^{***}$ \\
Qwen3-8B$\to$gpt-oss-120b & 19,605 & 27.2 & 55.5 & $+28.3^{***}$ & 40.3 & 26.3 & $-14.0^{***}$ \\
Qwen3-14B$\to$gpt-oss-20b & 17,160 & 14.7 & 48.8 & $+34.1^{***}$ & 50.8 & 34.2 & $-16.6^{***}$ \\
Qwen3-14B$\to$gpt-oss-120b & 17,160 & 18.8 & 51.1 & $+32.3^{***}$ & 50.5 & 32.3 & $-18.3^{***}$ \\
gpt-oss-20b$\to$gpt-oss-120b & 24,038 & 17.9 & 61.3 & $+43.4^{***}$ & 50.5 & 23.7 & $-26.7^{***}$ \\
\bottomrule
\end{tabular}}
\ifdim\wd0>\linewidth\resizebox{\linewidth}{!}{\usebox{0}}\else\usebox{0}\fi

\endgroup
\end{table}


\begin{table}[p]
\centering
\caption{Backtracking pilot. Reported pilot sample sizes and continuation behaviour. Category percentages refer to the 1,079 correct departures, not to all continuations. Explicit backtracking, change of direction, and restarting retain the category labels of the saved pilot summary. The manual agreement check reported 20/20 agreement; this is a small check, not validation of every continuation. This table describes the pilot and is separate from the full corrected rescue experiment.}
\label{tab:backtracking-pilot}
\begingroup
\small
\setlength{\tabcolsep}{3pt}
\renewcommand{\arraystretch}{1.12}
\sbox{0}{%
\begin{tabular}{@{}lr@{}}
\toprule
\textbf{Sample sizes} & Count \\
\midrule
All continuations & 9,529 \\
Correct continuations & 4,921 \\
Correct departures & 1,079 \\
\midrule
\multicolumn{2}{@{}l}{\textbf{Behaviour among the 1,079 correct departures}} \\
\addlinespace[2pt]
Behaviour & Percentage (\%) \\
\midrule
Explicit backtrack & 80.9 \\
Change direction & 18.1 \\
Restart & 1.0 \\
\bottomrule
\end{tabular}}
\ifdim\wd0>\linewidth\resizebox{\linewidth}{!}{\usebox{0}}\else\usebox{0}\fi

\endgroup
\end{table}


\clearpage
\section{Example injection prompts}

\begin{figure}[h]
    \centering
    \begin{tcolorbox}[colframe=mygrey, before upper={\raggedright\emergencystretch=2em}]
\allowbreak{}<|start|>\allowbreak{}system\allowbreak{}<|message|>\allowbreak{}You are ChatGPT, a large language model trained by OpenAI. \newline
Knowledge cutoff: 2024-06 \newline
Current date: 2025-12-20 \newline

Reasoning: medium \newline

\# Valid channels: analysis, commentary, final. Channel must be included for every message.\allowbreak{}<|end|>\allowbreak{}\allowbreak{}<|start|>\allowbreak{}developer\allowbreak{}<|message|>\allowbreak{}\# Instructions \newline

Solve the following problem. Please make sure that your response only consists of a single letter corresponding to the correct answer choice. Do not include anything else in your final response. \newline

\allowbreak{}<|end|>\allowbreak{}\allowbreak{}<|start|>\allowbreak{}user\allowbreak{}<|message|>\allowbreak{}Among the following exoplanets, which one has the highest density? \newline

a) An Earth-mass and Earth-radius planet. \newline
b) A planet with 2 Earth masses and a density of approximately 5.5 g/cm$^3$. \newline
c) A planet with the same composition as Earth but 5 times more massive than Earth. \newline
d) A planet with the same composition as Earth but half the mass of Earth. \newline

A. d \newline
B. a \newline
C. b \newline
D. c\allowbreak{}<|end|>\allowbreak{}\allowbreak{}<|start|>\allowbreak{}assistant
\end{tcolorbox}
    \caption{Example of a prompt formatted for the OpenAI gpt-oss family on question 1 of GPQA Diamond.}
    \label{fig:example_gpt_format}
\end{figure}

\begin{figure}[t]
    \centering
    \begin{tcolorbox}[colframe=mygrey, before upper={\raggedright\emergencystretch=2em}]
\allowbreak{}<|start|>\allowbreak{}system\allowbreak{}<|message|>\allowbreak{}You are ChatGPT, a large language model trained by OpenAI. \newline
Knowledge cutoff: 2024-06 \newline
Current date: 2025-12-20 \newline

Reasoning: medium \newline

\# Valid channels: analysis, commentary, final. Channel must be included for every message.\allowbreak{}<|end|>\allowbreak{}\allowbreak{}<|start|>\allowbreak{}developer\allowbreak{}<|message|>\allowbreak{}\# Instructions \newline

Solve the following problem. Please make sure that your response only consists of a single letter corresponding to the correct answer choice. Do not include anything else in your final response. \newline

\allowbreak{}<|end|>\allowbreak{}\allowbreak{}<|start|>\allowbreak{}user\allowbreak{}<|message|>\allowbreak{}Among the following exoplanets, which one has the highest density? \newline

a) An Earth-mass and Earth-radius planet. \newline
b) A planet with 2 Earth masses and a density of approximately 5.5 g/cm$^3$. \newline
c) A planet with the same composition as Earth but 5 times more massive than Earth. \newline
d) A planet with the same composition as Earth but half the mass of Earth. \newline

A. d \newline
B. a \newline
C. b \newline
D. c\allowbreak{}<|end|>\allowbreak{}\allowbreak{}<|start|>\allowbreak{}assistant\allowbreak{}<|channel|>\allowbreak{}analysis\allowbreak{}<|message|>\allowbreak{}\allowbreak{}<|end|>\allowbreak{}\allowbreak{}<|start|>\allowbreak{}assistant\allowbreak{}<|channel|>\allowbreak{}final\allowbreak{}<|message|>\allowbreak{}
\end{tcolorbox}
    \caption{Example of a non-reasoning prompt formatted for the OpenAI gpt-oss family on question 1 of GPQA Diamond.}
    \label{fig:example_gpt_no_reason_format}
\end{figure}

\begin{figure}[t]
    \centering
    \begin{tcolorbox}[colframe=mygrey, before upper={\raggedright\emergencystretch=2em}, fontupper=\scriptsize]
\allowbreak{}<|start|>\allowbreak{}system\allowbreak{}<|message|>\allowbreak{}You are ChatGPT, a large language model trained by OpenAI. \newline
Knowledge cutoff: 2024-06 \newline
Current date: 2025-12-20 \newline

Reasoning: medium \newline

\# Valid channels: analysis, commentary, final. Channel must be included for every message.\allowbreak{}<|end|>\allowbreak{}\allowbreak{}<|start|>\allowbreak{}developer\allowbreak{}<|message|>\allowbreak{}\# Instructions \newline

Solve the following problem. Please make sure that your response only consists of a single letter corresponding to the correct answer choice. Do not include anything else in your final response. \newline

\allowbreak{}<|end|>\allowbreak{}\allowbreak{}<|start|>\allowbreak{}user\allowbreak{}<|message|>\allowbreak{}Among the following exoplanets, which one has the highest density? \newline

a) An Earth-mass and Earth-radius planet. \newline
b) A planet with 2 Earth masses and a density of approximately 5.5 g/cm$^3$. \newline
c) A planet with the same composition as Earth but 5 times more massive than Earth. \newline
d) A planet with the same composition as Earth but half the mass of Earth. \newline

A. d \newline
B. a \newline
C. b \newline
D. c\allowbreak{}<|end|>\allowbreak{}\allowbreak{}<|start|>\allowbreak{}assistant\allowbreak{}<|channel|>\allowbreak{}analysis\allowbreak{}<|message|>\allowbreak{}We need to interpret question: Which exoplanet has the highest density among choices. Let's compute approximate densities.

\begin{itemize}
\item a: Earth-mass and Earth-radius planet. Density equal to Earth's: $\sim 5.5\,\mathrm{g/cm^3}$.

\item b: planet with $2$ Earth masses and density approx $5.5\,\mathrm{g/cm^3}$. So density equal to Earth's.

\item c: same composition as Earth but $5$ times more massive. For same composition, density increases? For rocks, as mass increases radius increases less than linear; for more mass, gravitational compression increases density. For composition same, radius increases, but compressional effect leads to slightly higher density. Densities: For planets of larger mass, density increases slightly but not huge. Earth-like composition typical density $\sim 5.5\,\mathrm{g/cm^3}$. For $5M_{\oplus}$, typical density maybe around $6$--$7\,\mathrm{g/cc}$? But we need relative to others.

\item d: same composition but half mass. Lower density due to less compression? Might be similar or slightly lower.
\end{itemize}

Need to select who has highest density. Option c likely highest due to greater gravitational compression.

Let's analyze more quantitatively? For exoplanet mass scaling: For planets dominated by iron/rock, radius scales roughly $M^{0.27}$? Actually radius scaling $\sim M^{0.27}$ for Earth-like composition: $R \propto M^{0.27}$ (see equation). Then density $\rho = M/(4/3\pi R^3) \propto M / M^{0.81} = M^{0.19}$. So density scales as $M^{0.19}$. For $M=5M_{\oplus}$, density ratio $5^{0.19} \approx 1.3$. So density $\sim 1.3\times$ Earth's density $\sim 7.1\,\mathrm{g/cc}$. That's higher than Earth's $5.5\,\mathrm{g/cm^3}$. So option c has highest.

Option b has same density as Earth, $5.5$. Option d probably lower maybe $5.5/1.3$? Wait $M=0.5 \rightarrow$ density ratio $0.5^{0.19}\approx 0.86 \rightarrow$ density $\sim 4.7\,\mathrm{g/cc}$.

Thus c highest. Option c is letter? Options list: a,b,c,d labeled. The answer choices: A.\ d B.\ a C.\ b D.\ c. So the correct answer is D? Wait mapping: options mapping: They wrote first list a,b,c,d as exoplanets. Then answer list A,B,C,D corresponding to d,a,b,c. So we need to choose which answer letter corresponds to the right explanation? Let's see: They ask ``Which one has highest density?'' Options a,b,c,d. Then answer list A.\ d B.\ a C.\ b D.\ c. So we choose which letter: the correct exoplanet is c. So answer letter D (since D corresponds to c). So output ``D''.

But double-check: Could Option b have same density but we have to pick highest density. Option c has higher density, so correct.

Thus final answer letter: D\allowbreak{}<|end|>\allowbreak{}\allowbreak{}<|start|>\allowbreak{}assistant\allowbreak{}<|channel|>\allowbreak{}final\allowbreak{}<|message|>\allowbreak{}D
\end{tcolorbox}
    \caption{Example of a full response (from run 1) of OpenAI gpt-oss-20b on question 1 of GPQA Diamond.}
    \label{fig:example_gpt_full}
\end{figure}

\begin{figure}[t]
    \centering
    \begin{tcolorbox}[colframe=mygrey, before upper={\raggedright\emergencystretch=2em}]
\allowbreak{}<|start|>\allowbreak{}system\allowbreak{}<|message|>\allowbreak{}You are ChatGPT, a large language model trained by OpenAI. \newline
Knowledge cutoff: 2024-06 \newline
Current date: 2025-12-20 \newline

Reasoning: medium \newline

\# Valid channels: analysis, commentary, final. Channel must be included for every message.\allowbreak{}<|end|>\allowbreak{}\allowbreak{}<|start|>\allowbreak{}developer\allowbreak{}<|message|>\allowbreak{}\# Instructions \newline

Solve the following problem. Please make sure that your response only consists of a single letter corresponding to the correct answer choice. Do not include anything else in your final response. \newline

\allowbreak{}<|end|>\allowbreak{}\allowbreak{}<|start|>\allowbreak{}user\allowbreak{}<|message|>\allowbreak{}Among the following exoplanets, which one has the highest density? \newline

a) An Earth-mass and Earth-radius planet. \newline
b) A planet with 2 Earth masses and a density of approximately 5.5 g/cm$^3$. \newline
c) A planet with the same composition as Earth but 5 times more massive than Earth. \newline
d) A planet with the same composition as Earth but half the mass of Earth. \newline

A. d \newline
B. a \newline
C. b \newline
D. c\allowbreak{}<|end|>\allowbreak{}\allowbreak{}<|start|>\allowbreak{}assistant\allowbreak{}<|channel|>\allowbreak{}analysis\allowbreak{}<|message|>\allowbreak{}We need to interpret question: Which exoplanet has the highest density among choices. Let's compute approximate densities.

\begin{itemize}
\item a: Earth-mass and Earth-radius planet. Density equal to Earth's: $\sim 5.5\,\mathrm{g/cm^3}$.

\item b: planet with $2$ Earth masses and density approx $5$\allowbreak{}<|end|>\allowbreak{}\allowbreak{}<|start|>\allowbreak{}assistant\allowbreak{}<|channel|>\allowbreak{}final\allowbreak{}<|message|>\allowbreak{}\end{itemize}
\end{tcolorbox}
    \caption{Example of a 10\% reasoning decile (from run 1) of OpenAI gpt-oss-20b on question 1 of GPQA Diamond.}
    \label{fig:fig:example_gpt_10}
\end{figure}

\begin{figure}[t]
    \centering
    \begin{tcolorbox}[colframe=mygrey, before upper={\raggedright\emergencystretch=2em}]
\allowbreak{}<|im\_start|>\allowbreak{}system \newline
Solve the following problem. Please make sure that your response only consists of a single letter corresponding to the correct answer choice. Do not include anything else in your final response.\allowbreak{}<|im\_end|>\allowbreak{} \newline
\allowbreak{}<|im\_start|>\allowbreak{}user \newline
Among the following exoplanets, which one has the highest density? \newline

a) An Earth-mass and Earth-radius planet. \newline
b) A planet with 2 Earth masses and a density of approximately 5.5 g/cm\^{}3. \newline
c) A planet with the same composition as Earth but 5 times more massive than Earth. \newline
d) A planet with the same composition as Earth but half the mass of Earth. \newline

A. d \newline
B. a \newline
C. b \newline
D. c\allowbreak{}<|im\_end|>\allowbreak{} \newline
\allowbreak{}<|im\_start|>\allowbreak{}assistant
\end{tcolorbox}
    \caption{Example of a prompt formatted for the Qwen3-4B model on question 1 of GPQA Diamond.}
    \label{fig:example_qwen_format}
\end{figure}

\begin{figure}[t]
    \centering
    \begin{tcolorbox}[colframe=mygrey, before upper={\raggedright\emergencystretch=2em}]
\allowbreak{}<|im\_start|>\allowbreak{}system \newline
Solve the following problem. Please make sure that your response only consists of a single letter corresponding to the correct answer choice. Do not include anything else in your final response.\allowbreak{}<|im\_end|>\allowbreak{} \newline
\allowbreak{}<|im\_start|>\allowbreak{}user \newline
Among the following exoplanets, which one has the highest density? \newline

a) An Earth-mass and Earth-radius planet. \newline
b) A planet with 2 Earth masses and a density of approximately 5.5 g/cm\^{}3. \newline
c) A planet with the same composition as Earth but 5 times more massive than Earth. \newline
d) A planet with the same composition as Earth but half the mass of Earth. \newline

A. d \newline
B. a \newline
C. b \newline
D. c\allowbreak{}<|im\_end|>\allowbreak{} \newline
\allowbreak{}<|im\_start|>\allowbreak{}assistant \newline
\allowbreak{}<think>\allowbreak{} \newline
\newline
\allowbreak{}</think>\allowbreak{} \newline
\end{tcolorbox}
    \caption{Example of a non-reasoning prompt formatted for the Qwen3-4B model on question 1 of GPQA Diamond.}
    \label{fig:example_qwen_no_reason_format}
\end{figure}

\begin{figure}[t]
    \centering
    \begin{tcolorbox}[colframe=mygrey, fontupper=\scriptsize, before upper={\raggedright\emergencystretch=2em}]
\allowbreak{}<|im\_start|>\allowbreak{}system \newline
Solve the following problem. Please make sure that your response only consists of a single letter corresponding to the correct answer choice. Do not include anything else in your final response.\allowbreak{}<|im\_end|>\allowbreak{} \newline
\allowbreak{}<|im\_start|>\allowbreak{}user \newline
Among the following exoplanets, which one has the highest density? \newline

a) An Earth-mass and Earth-radius planet. \newline
b) A planet with 2 Earth masses and a density of approximately 5.5 g/cm\^{}3. \newline
c) A planet with the same composition as Earth but 5 times more massive than Earth. \newline
d) A planet with the same composition as Earth but half the mass of Earth. \newline

A. d \newline
B. a \newline
C. b \newline
D. c\allowbreak{}<|im\_end|>\allowbreak{} \newline
\allowbreak{}<|im\_start|>\allowbreak{}assistant \newline
\allowbreak{}<think>\allowbreak{} \newline
Okay, let's try to figure out which exoplanet has the highest density. The question gives me four options, and I need to determine which one has the highest density. Let me start by recalling what density is. Density is mass divided by volume, right? So, the formula is density = mass/volume. \newline

Now, the problem is about exoplanets, so I need to think about how mass and radius relate to density. For planets, if they have the same composition, like Earth, their density would depend on their mass and radius. But if the mass changes, the radius might change as well, depending on how the planet is structured. \newline

Let me look at each option one by one. \newline

Option a) An Earth-mass and Earth-radius planet. So, this is just Earth. Earth's density is about 5.5 g/cm$^3$. \newline

Option b) A planet with 2 Earth masses and a density of approximately 5.5 g/cm$^3$. So, this is a planet that is twice as massive as Earth but has the same density as Earth. \newline

Option c) A planet with the same composition as Earth but 5 times more massive than Earth. So, if composition is same as Earth, then the density would depend on how the mass affects the radius. \newline

Option d) A planet with same composition, half mass. So same density as Earth. \newline

[...] \newline

But the question is likely expecting a simple comparison. Since option c is same composition as Earth, but 5 times more massive, the density would be higher than Earth. Therefore, the answer is c. \newline

But earlier reasoning said that if same composition, density is same. But that is only if the planet is scaled up, leading to the same density. But if the planet is more compressed, the density would be higher. The problem is, the question doesn't give enough information about the planet's structure. \newline

But given that the question is from a test, it's likely that the intended answer is option c. Because for a planet with same composition, more mass implies higher density, since it's more compressed. Therefore, option c has higher density than the others. \newline

So, in that case, the answer is option c, which is D. \newline
\allowbreak{}</think>\allowbreak{} \newline

D
\end{tcolorbox}
    \caption{Example of a full response (from run 1) of Qwen3-4B on question 1 of GPQA Diamond. The reasoning trace has been shortened with [...] for display purposes (full trace: 7,639 tokens).}
    \label{fig:example_qwen_full}
\end{figure}

\begin{figure}[t]
    \centering
    \begin{tcolorbox}[colframe=mygrey, fontupper=\scriptsize, before upper={\raggedright\emergencystretch=2em}]
\allowbreak{}<|im\_start|>\allowbreak{}system \newline
Solve the following problem. Please make sure that your response only consists of a single letter corresponding to the correct answer choice. Do not include anything else in your final response.\allowbreak{}<|im\_end|>\allowbreak{} \newline
\allowbreak{}<|im\_start|>\allowbreak{}user \newline
Among the following exoplanets, which one has the highest density? \newline

a) An Earth-mass and Earth-radius planet. \newline
b) A planet with 2 Earth masses and a density of approximately 5.5 g/cm\^{}3. \newline
c) A planet with the same composition as Earth but 5 times more massive than Earth. \newline
d) A planet with the same composition as Earth but half the mass of Earth. \newline

A. d \newline
B. a \newline
C. b \newline
D. c\allowbreak{}<|im\_end|>\allowbreak{} \newline
\allowbreak{}<|im\_start|>\allowbreak{}assistant \newline
\allowbreak{}<think>\allowbreak{} \newline
Okay, let's try to figure out which exoplanet has the highest density. The question gives me four options, and I need to determine which one has the highest density. Let me start by recalling what density is. Density is mass divided by volume, right? So, the formula is density = mass/volume. \newline

Now, the problem is about exoplanets, so I need to think about how mass and radius relate to density. For planets, if they have the same composition, like Earth, their density would depend on their mass and radius. But if the mass changes, the radius might change as well, depending on how the planet is structured. For example, if a planet is more massive, it might be more compressed, but that's not always the case. Wait, but if they have the same composition, then maybe the density is more directly related to mass and radius. \newline

Let me look at each option one by one. \newline

Option a) An Earth-mass and Earth-radius planet. So, this is just Earth. Earth's density is about 5.5 g/cm$^3$. But the question is, if this is the same as Earth, then its density is 5.5. But the problem is whether this is the same as Earth's density. Well, Earth's density is approximately 5.5 g/cm$^3$. So maybe this is just Earth's density. \newline

Option b) A planet with 2 Earth masses and a density of approximately 5.5 g/cm$^3$. So, this is a planet that is twice as massive as Earth but has the same density as Earth. Wait, if density is mass/volume, then if the density is the same, then the volume must be proportional to the mass. So, if it's twice as massive, then the volume would be twice as much. But then the radius would be... Well, volume of a sphere is (4/3)$\pi$r$^3$. So, if the volume is twice as much, then r would be cube root of 2 times Earth's radius. So that would make the radius larger than Earth's. So this planet would be more massive but with a larger radius than Earth, but with the same density. So in this case, the density is 5.5, same as Earth. But the actual density is given as 5.5 here. Wait, but maybe this is a planet that's not composed of the same material as Earth, but just has a density of 5.5. But why is that? Because if the planet is more massive, but same density, then it's volume is larger, so radius is larger. But the point is, for this option, the density is given as approximately 5.5. So maybe this is a planet with same density as Earth but more mass. \newline

Option c) A planet with the same composition as Earth but 5 times more massive than Earth. So, if composition is same as Earth, then the density would depend on how the mass affects the radius. If the planet is more massive, but same composition, then it would be more compact. Wait, but Earth's density is 5.5 g/cm$^3$. If a planet is more massive but same composition, then how does that affect its radius? Wait, if the mass increases, but the density remains the same, then the radius would increase. But if the density is same, then mass is proportional to volume, so radius would increase with the cube root of mass. But if the mass is 5 times that of Earth, then radius would be cube root of 5 times Earth's radius. But if the composition is the same, then maybe the radius would be larger. However, the density would still be \newline
\newline
Considering the limited time by the user, I have to give the solution based on the thinking directly now. \newline
\allowbreak{}</think>\allowbreak{} \newline
\end{tcolorbox}
    \caption{Example of a 10\% reasoning decile (from run 1) of Qwen3-4B on question 1 of GPQA Diamond.}
    \label{fig:example_qwen_10}
\end{figure}

\begin{figure}[t]
    \centering
    \begin{tcolorbox}[colframe=mygrey, fontupper=\scriptsize, before upper={\raggedright\emergencystretch=2em}]
\allowbreak{}<|im\_start|>\allowbreak{}system \newline
Solve the following problem. Please make sure that your response only consists of a single letter corresponding to the correct answer choice. Do not include anything else in your final response.\allowbreak{}<|im\_end|>\allowbreak{} \newline
\allowbreak{}<|im\_start|>\allowbreak{}user \newline
Among the following exoplanets, which one has the highest density? \newline

a) An Earth-mass and Earth-radius planet. \newline
b) A planet with 2 Earth masses and a density of approximately 5.5 g/cm\^{}3. \newline
c) A planet with the same composition as Earth but 5 times more massive than Earth. \newline
d) A planet with the same composition as Earth but half the mass of Earth. \newline

A. d \newline
B. a \newline
C. b \newline
D. c\allowbreak{}<|im\_end|>\allowbreak{} \newline
\allowbreak{}<|im\_start|>\allowbreak{}assistant \newline
\allowbreak{}<think>\allowbreak{} \newline
Okay, let's try to figure out which exoplanet has the highest density. The question gives me four options, and I need to determine which one has the highest density. Let me start by recalling what density is. Density is mass divided by volume, right? So, the formula is density = mass/volume. \newline

Now, the problem is about exoplanets, so I need to think about how mass and radius relate to density. For planets, if they have the same composition, like Earth, their density would depend on their mass and radius. But if the mass changes, the radius might change as well, depending on how the planet is structured. For example, if a planet is more massive, it might be more compressed, but that's not always the case. Wait, but if they have the same composition, then maybe the density is more directly related to mass and radius. \newline

Let me look at each option one by one. \newline

Option a) An Earth-mass and Earth-radius planet. So, this is just Earth. Earth's density is about 5.5 g/cm$^3$. But the question is, if this is the same as Earth, then its density is 5.5. But the problem is whether this is the same as Earth's density. Well, Earth's density is approximately 5.5 g/cm$^3$. So maybe this is just Earth's density. \newline

Option b) A planet with 2 Earth masses and a density of approximately 5.5 g/cm$^3$. So, this is a planet that is twice as massive as Earth but has the same density as Earth. Wait, if density is mass/volume, then if the density is the same, then the volume must be proportional to the mass. So, if it's twice as massive, then the volume would be twice as much. But then the radius would be... Well, volume of a sphere is (4/3)$\pi$r$^3$. So, if the volume is twice as much, then r would be cube root of 2 times Earth's radius. So that would make the radius larger than Earth's. So this planet would be more massive but with a larger radius than Earth, but with the same density. So in this case, the density is 5.5, same as Earth. But the actual density is given as 5.5 here. Wait, but maybe this is a planet that's not composed of the same material as Earth, but just has a density of 5.5. But why is that? Because if the planet is more massive, but same density, then it's volume is larger, so radius is larger. But the point is, for this option, the density is given as approximately 5.5. So maybe this is a planet with same density as Earth but more mass. \newline

Option c) A planet with the same composition as Earth but 5 times more massive than Earth. So, if composition is same as Earth, then the density would depend on how the mass affects the radius. If the planet is more massive, but same composition, then it would be more compact. Wait, but Earth's density is 5.5 g/cm$^3$. If a planet is more massive but same composition, then how does that affect its radius? Wait, if the mass increases, but the density remains the same, then the radius would increase. But if the density is same, then mass is proportional to volume, so radius would increase with the cube root of mass. But if the mass is 5 times that of Earth, then radius would be cube root of 5 times Earth's radius. But if the composition is the same, then maybe the radius would be larger. However, the density would still be \newline
\allowbreak{}</think>\allowbreak{} \newline
\end{tcolorbox}
    \caption{Example of a 10\% reasoning decile with alternative early-stopping suffix.}
    \label{fig:example_qwen_10_alternative}
\end{figure}

\end{document}